\documentclass[10pt,journal,compsoc]{IEEEtran}

\ifCLASSOPTIONcompsoc
  \usepackage[nocompress]{cite}
\else
  \usepackage{cite}
\fi

\newcommand\MYhyperrefoptions{bookmarks=true,bookmarksnumbered=true,
pdfpagemode={UseOutlines},plainpages=false,pdfpagelabels=true,
colorlinks=true,linkcolor={black},citecolor={black},urlcolor={black},
pdftitle={Multispectral Biometrics System Framework: Application to Presentation Attack Detection},
pdfsubject={Biometrics},
pdfauthor={Leonidas Spinoulas},
pdfkeywords={Multispectral Biometrics, Biometrics System Framework, Multispectral Cameras, Presentation Attack Detection.}}

\ifCLASSINFOpdf
\usepackage[\MYhyperrefoptions,pdftex]{hyperref}
\else
\usepackage[\MYhyperrefoptions,breaklinks=true,dvips]{hyperref}
\usepackage{breakurl}
\fi

\ifCLASSOPTIONcompsoc
  \usepackage[caption=false,font=footnotesize,labelfont=sf,textfont=sf]{subfig}
\else
 \usepackage[caption=false,font=footnotesize]{subfig}
\fi

\usepackage{booktabs}
\usepackage{graphicx}
\usepackage{multirow}
\usepackage{multicol}
\usepackage{balance}
\usepackage{color, colortbl, soul}
\definecolor{Gray}{gray}{0.95}

\newif\ifarxiv
\arxivtrue 

\begin{document}
\title{Multispectral Biometrics System Framework: Application to Presentation Attack Detection}
%
%
%
%

\ifarxiv
\author{Leonidas~Spinoulas,
        Mohamed~Hussein,~\IEEEmembership{Member,~IEEE},
        David~Geissb{\"u}hler,
        Joe~Mathai, Oswin~G.~Almeida, Guillaume~Clivaz, S{\'e}bastien~Marcel, 
        and~Wael~AbdAlmageed,~\IEEEmembership{Member,~IEEE}
\IEEEcompsocitemizethanks{\IEEEcompsocthanksitem L. Spinoulas, M. Hussein, J. Mathai, O. G. Almeida and W. AbdAlmageed are with the Information Sciences Institute (University of Southern California), Marina Del Rey, CA, 90292.\protect\\
M. Hussein is also with the Faculty of Engineering, Alexandria University, Alexandria, Egypt 21544.\protect\\
E-mail: lspinoulas@isi.edu, mehussein@isi.edu, jmathai@isi.edu, wamageed@isi.edu
\IEEEcompsocthanksitem D. Geissb{\"u}hler, G. Clivaz and S. Marcel are
with the Idiap Research Institute, CH-1920 Martigny.}}
\else
\author{Leonidas~Spinoulas,
        Mohamed~Hussein,~\IEEEmembership{Member,~IEEE},
        David~Geissb{\"u}hler,
        Joe~Mathai, Oswin~G.~Almeida, Guillaume~Clivaz, S{\'e}bastien~Marcel, 
        and~Wael~AbdAlmageed,~\IEEEmembership{Member,~IEEE}
\IEEEcompsocitemizethanks{\IEEEcompsocthanksitem L. Spinoulas, M. Hussein, J. Mathai, O. G. Almeida and W. AbdAlmageed are with the Information Sciences Institute (University of Southern California), Marina Del Rey, CA, 90292.\protect\\
M. Hussein is also with the Faculty of Engineering, Alexandria University, Alexandria, Egypt 21544.\protect\\
E-mail: lspinoulas@isi.edu, mehussein@isi.edu, jmathai@isi.edu, wamageed@isi.edu
\IEEEcompsocthanksitem D. Geissb{\"u}hler, G. Clivaz and S. Marcel are
with the Idiap Research Institute, CH-1920 Martigny.}
\thanks{Manuscript received June XX, 2020; revised XXXX XX, 2020.}}
\fi
%
%

\ifarxiv
\else
\markboth{IEEE Transactions on Biometrics, Behavior, and Identity Science,~Vol.~XX, No.~XX, XXXX~2020}%
{Multispectral Biometrics System Framework: Application to Presentation Attack Detection}
\fi
%



\IEEEtitleabstractindextext{%
\begin{abstract}
In this work, we present a general framework for building a biometrics system capable of capturing multispectral data from a series of sensors synchronized with active illumination sources. The framework unifies the system design for different biometric modalities and its realization on face, finger and iris data is described in detail. To the best of our knowledge, the presented design is the first to employ such a diverse set of electromagnetic spectrum bands, ranging from visible to long-wave-infrared wavelengths, and is capable of acquiring large volumes of data in seconds. Having performed a series of data collections, we run a comprehensive analysis on the captured data using a deep-learning classifier for presentation attack detection. Our study follows a data-centric approach attempting to highlight the strengths and weaknesses of each spectral band at distinguishing live from fake samples.
\end{abstract}

}

\maketitle

\IEEEdisplaynontitleabstractindextext

\ifCLASSOPTIONpeerreview
\begin{center} \bfseries EDICS Category: 3-BBND \end{center}
\fi
%
\IEEEpeerreviewmaketitle

\ifCLASSOPTIONcompsoc
\IEEEraisesectionheading{\section{Introduction}\label{sec:introduction}}
\else
\section{Introduction}
\label{sec:introduction}
\fi

%
%
%
%
\IEEEPARstart{B}{iometric} sensors have become ubiquitous in recent years with ever more increasing industries introducing some form of biometric authentication for enhancing security or simplifying user interaction. They can be found on everyday items such as smartphones and laptops as well as in facilities requiring high levels of security such as banks, airports or border control. Even though the wide usability of biometric sensors is intended to enhance security, it also comes with a risk of increased spoofing attempts. At the same time, the large availability of commercial sensors enables access to the underlying technology for testing various approaches aiming at concealing one's identity or impersonating someone else, which is the definition of a \emph{Presentation Attack} (PA). Besides, advances in materials technology have already enabled the development of \emph{Presentation Attack Instruments} (PAIs) capable at successfully spoofing existing biometric systems~\cite{Spurny2015, Lafkih2015, ba_under_threat}.

\emph{Presentation Attack Detection} (PAD) has attracted a lot of interest with a long list of publications focusing on devising algorithms where data from existing biometric sensors are used~\cite{Marcel2019}. In this work, we approach the PAD problem from a sensory perspective and attempt to design a system that relies mostly on the captured data which should ideally exhibit a distinctive response for PAIs. We focus on capturing spectral data, which refers to the acquisition of images of various bands of the electromagnetic spectrum for extracting additional information of an object beyond its visible spectrum~\cite{Munir2019}. The higher dimensionality of multispectral data enables detection of other than skin materials based on their spectral characteristics~\cite{Roui-Abidi2009}. A comprehensive analysis of the spectral emission of skin and different fake materials~\cite{Steiner2016} shows that in higher than visible wavelengths, the remission properties of skin converges for different skin types (i.e., different race or ethnicity) compared to a diverse set of lifeless substances. Additionally, multispectral data offer a series of advantages over conventional visible light imaging, including visibility through occlusions as well as being unaffected by ambient illumination conditions. 

Spectral imaging has been studied for over a decade with applications to medical imaging, food engineering, remote sensing, industrial applications and security~\cite{Sigrononi2019}. However, its use in biometrics is still at its infancy. A few prototype systems can be found in the literature for \emph{face}~\cite{Steiner2016}, \emph{finger}~\cite{Engelsma2019} and \emph{iris}~\cite{Venkatesh2019} data but usually employ a small set of wavelengths~\cite{Zhang2016}. Commercial sensors are still very limited (e.g.,~\cite{realsense, lumidigm, vistaey2}) and mainly use a few wavelengths in the visible (VIS) or near-infrared (NIR) spectra. Lately, hybrid sensors have also appeared on smartphones combining VIS, NIR and depth measurements. The majority of existing PAD literature on multispectral data has relied on such commercial sensors for studying PAD paradigms (e.g., ~\cite{Chingovska2016, Raghavendra2017, Agarwal2017}).

In general, systems capturing spectral data can be grouped in $4$ main categories:

\begin{figure*}[!t]
    \centering
    \includegraphics{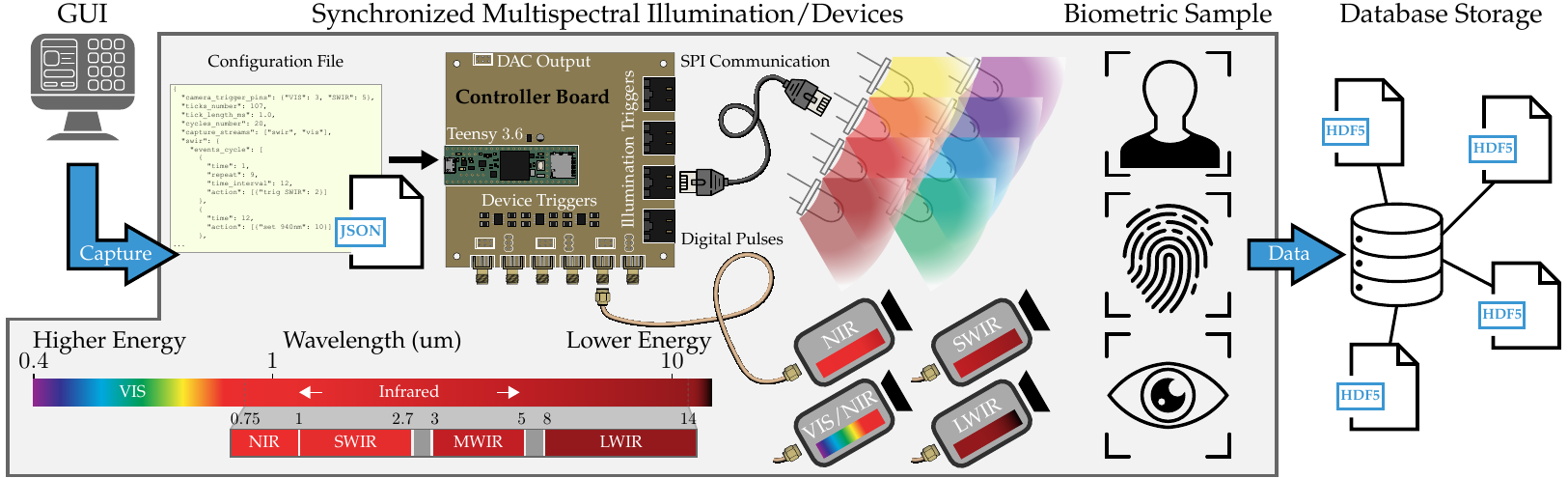}
    \caption{Main components of the proposed multispectral biometrics system framework: A biometric sample is observed by a sensor suite comprised of various multispectral data capture devices. A set of multispectral illumination sources is synchronized with the sensors through an electronic controller board. A computer provides the synchronization sequence through a JSON file and sends capture commands that bring the controller and sensors into a capture loop leading to a sequence of synchronized multispectral data from all devices. All captured data is then packaged into an HDF5 file and sent to a database for storage and further processing.}
    \label{fig:system}
\end{figure*}

\begin{enumerate}
    \item Multispectral image acquisition using multiple cameras inherently sensitive at different wavelength regimes or employing band-pass filters~\cite{Hussein2018}.
    \item Hyperspectral imagers~\cite{Agnieszka2018}.
    \item Single cameras performing sequential image acquisition with a rotating wheel of band-pass filters~\cite{Brauers2008}.
    \item Single cameras with Bayer-like band-pass filter patterns~\cite{Wu2020}.
\end{enumerate}

In this work, we follow the first approach and propose a unified framework for multispectral biometric data capture, by combining a variety of cameras, synchronized with a set of illumination sources for collecting data at different sub-bands of the VIS, NIR, short-wave-infrared (SWIR) and long-wave-infrared (LWIR) spectra. To the best of our knowledge this is the first work to enable capture of such diverse types of multispectral data in a unified system framework for different biometric modalities. In the following sections, we analyze the proposed system framework and emphasize on the merits of the captured data at detecting PAs.

\section{System Framework and Design}
\label{sec:system_design}

In this section, we analyze the proposed multispectral biometrics system framework and present its realization with three different biometric sensor suites applied on \emph{face}, \emph{finger} and \emph{iris} biometric data, respectively.

The main concept of our framework is presented in Fig.~\ref{fig:system} and is initially described here at a very high level. A biometric sample is observed by a sensor suite comprised of various multispectral data capture devices. A set of multispectral illumination sources is synchronized with the sensors through an electronic controller board. A computer uses a Graphical User Interface (GUI) which provides the synchronization sequence through a JSON configuration file and sends capture commands that bring the controller and sensors into a capture loop leading to a sequence of synchronized multispectral frames from all devices. All captured data is then packaged into an HDF5 file and sent to a database for storage and processing. The computer also provides preview capabilities to the user by streaming data in real-time from each device while the GUI is in operation.

Our system design (both in terms of hardware and software) is governed by four key principles:

\begin{enumerate}
    \item \textbf{\underline{Flexibility}}: Illumination sources and capture devices can be easily replaced with alternate ones with no or minimal effort both in terms of hardware and software development.
    \item \textbf{\underline{Modularity}}: Whole components of the system can be disabled or removed without affecting the overall system's functionality by simply modifying the JSON configuration file.
    \item \textbf{\underline{Legacy compatibility}}: The system must provide at least some type of data that can be used for biometric identification through matching with data from older sensors and biometric templates available in existing databases. 
    \item \textbf{\underline{Complementarity}}: The variety of capture devices and illumination sources used aim at providing complementary information about the biometric sample aiding the underlying task at hand.
\end{enumerate}
We first describe all the common components of our system (as depicted in Fig.~\ref{fig:system}) and then discuss the specifics of each sensor suite for the three studied biometric modalities.

\begin{figure}[!t]
    \centering
    \subfloat[LED illumination module, controlled through an LED driver~\cite{led_driver}. Contains $16$ slots for SMD LEDs.]{\includegraphics[scale=0.9]{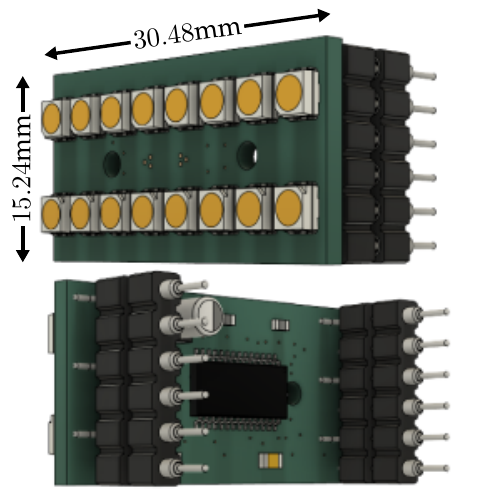}}
    \hspace{0.5cm}
    \subfloat[Teensy $3.6$ microcontroller~\cite{teensy} used in the controller board of Fig.~\ref{fig:system}.]{\includegraphics[scale=0.9]{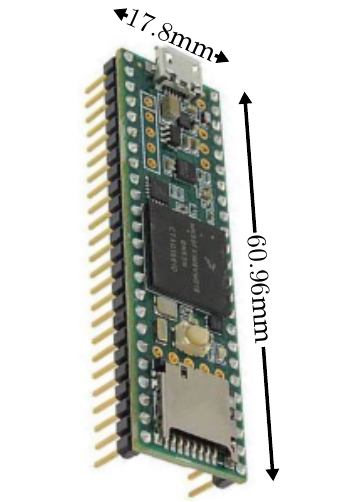}}
    \caption{System's main electronic components.}
    \label{fig:electronic_components}
\end{figure}

\subsection{Hardware}
\label{subsec:hardware}
The hardware design follows all principles described above providing a versatile system which can be easily customized for different application needs. \vspace{0.1cm}

\begin{table*}[!t]
    \caption{Summary of cameras used in all presented biometric sensor suites along with their main specifications.}
    \label{tab:list_of_cameras}
    \centering
    \resizebox{\textwidth}{!}{
    \begin{tabular}{p{1cm}p{0.1cm}||c|cc|cc|c|cc|ccc}
         \multicolumn{2}{c||}{\textbf{Regime}} & \textbf{RGB} & \multicolumn{2}{c|}{\textbf{NIR}} & \multicolumn{2}{c|}{\textbf{VIS/NIR}} & \textbf{SWIR} & \multicolumn{2}{c|}{\textbf{Thermal}} & \multicolumn{3}{c}{\textbf{RGB/NIR/Depth}} \\ \toprule \hline
         
         \multicolumn{2}{c||}{\multirow{3}{*}{\textbf{Product}}} & Basler~\cite{basler_rgb}, & Basler~\cite{basler_nir_face},& Basler~\cite{basler_nir_iris},& Basler~\cite{basler_nir_finger},& IrisID~\cite{iris_id},& Xenics~\cite{bobcat},& FLIR~\cite{boson_face},& FLIR~\cite{boson_iris},& \multicolumn{3}{c}{Intel~\cite{realsense},}\\
         
         &  & acA1920 &  acA1920 & acA4096 & acA1300  & iCam & Bobcat  & Boson & Boson &  \multicolumn{3}{c}{RealSense D435} \\
         & & -150uc & -150um & -30um & -60gmNIR & -7000 & 320-100 & 320-24$^{\circ}$ & 640-18$^{\circ}$&  -RGB & -NIR ($\times 2$) & -Depth \\ \hline
         
         \multicolumn{2}{c||}{\textbf{Comm.}} & \multirow{2}{*}{USB3} & \multirow{2}{*}{USB3} & \multirow{2}{*}{USB3} & \multirow{2}{*}{GigE} & \multirow{2}{*}{GigE} & \multirow{2}{*}{GigE} & \multirow{2}{*}{USB2} & \multirow{2}{*}{USB2} & \multicolumn{3}{c}{\multirow{2}{*}{USB3}}\\
         \multicolumn{2}{c||}{\textbf{Protocol}} & & & & & & & & & & &\\ \hline
         
         \multicolumn{2}{c||}{\textbf{Full Image}} &  $1984 \times$ & $1984 \times$ & $4096 \times$ & $1282 \times$ & $640 \times$ & $320 \times$ & $320 \times$ & $640 \times$  & $1280 \times$ & $1280 \times$ & $1280 \times$ \\
         \multicolumn{2}{c||}{\textbf{Size} (pixels)} & $1264\times 3$ & $1264$ &  $2168$ & $1026$ & $480 \times 2$ &$256$ & $256$ & $512$ &  $720\times3$ & $720$ & $720$  \\ \hline
         
         \multicolumn{2}{c||}{\textbf{Max Bit Depth}} & $8\times3$ bits& $10$ bits & $12$ bits & $12$ bits  &
         $8$ bits &
         $16$ bits & $16$ bits & $16$ bits &  $8\times3$ bits & $8$ bits & $16$ bits\\ \hline
         
         \multicolumn{2}{c||}{\textbf{Max}} & \multirow{2}{*}{$150$ fps} & \multirow{2}{*}{$150$ fps} & \multirow{2}{*}{$32$ fps} & \multirow{2}{*}{$60$ fps} &
         \multirow{2}{*}{N/A} &
         \multirow{2}{*}{$100$ fps} & \multirow{2}{*}{$60$ fps} & \multirow{2}{*}{$60$ fps} &  \multirow{2}{*}{$30$ fps} & \multirow{2}{*}{$90$ fps} & \multirow{2}{*}{$90$ fps} \\ 
         \multicolumn{2}{c||}{\textbf{Frame Rate}} & & & & & & & & & & & \\ \hline
         
         \multicolumn{2}{c||}{\textbf{Trigger}} & Hardware & Hardware & Hardware & Hardware & Software & Hardware & Software & Software & \multicolumn{3}{c}{Software} \\ \hline 
         
         \multicolumn{2}{c||}{\textbf{Lens Used}} & \cite{kowa_lens_face} & \cite{kowa_lens_face} & \cite{kowa_lens_iris} & \cite{eo_lens_finger} & Built-in & \cite{swir_lens_face} / \cite{swir_lens_finger} & Built-in & Built-in & \multicolumn{3}{c}{Built-in} \\ \hline
         
         \multicolumn{2}{c||}{\textbf{External Filter}} & No & Yes~\cite{nir_filter_face} & Yes~\cite{nir_filter_iris} & No & No & No & No & No & \multicolumn{3}{c}{No} \\ \hline

         \multirow{2}{*}{\textbf{Spectrum}} & Min & $400$nm & $650$nm & $650$nm & $400$nm & $400$nm & $900$nm & $8$um & $8$um &  $400$nm & $400$nm & \multirow{2}{*}{N/A}\\
         & Max & $700$nm & $1050$nm & $1050$nm & $1050$nm & $1050$nm & $1700$nm & $14$um & $14$um &  $700$nm & $1050$nm & \\ \hline 
         
         \multicolumn{2}{c||}{\multirow{2}{*}{\textbf{Sensor Suite}}} & \multirow{2}{*}{Face} & \multirow{2}{*}{Face ($\times 2$)} & \multirow{2}{*}{Iris} & \multirow{2}{*}{Finger} & \multirow{2}{*}{Iris} & Face / & \multirow{2}{*}{Face} & \multirow{2}{*}{Iris} & \multicolumn{3}{c}{\multirow{2}{*}{Face}} \\
          & & & & & & & Finger & & & & &\\
         \hline \bottomrule
    \end{tabular}
    }
\end{table*}

\paragraph*{\textbf{Illumination Modules}} We have designed a Light-Emitting Diode (LED) based illumination module which can be used as a building block for creating a larger array of LEDs in various spatial configurations. It is especially made for supporting Surface-Mount Device (SMD) LEDs for compactness. The module, shown in Fig.~\ref{fig:electronic_components}(a), contains $16$ slots for mounting LEDs and uses a Serial Peripheral Interface (SPI) communication LED driver chip~\cite{led_driver} which allows independent control of the current and Pulse-Width-Modulation (PWM) for each slot. LEDs can be turned on/off or their intensity can be modified using a sequence of bits. Since current is independently controlled for each position, it allows combining LEDs with different operating limits. \vspace{0.1cm}

\paragraph*{\textbf{Controller Board}} The controller board also follows a custom design and uses an Arduino-based  microcontroller (Teensy $3.6$~\cite{teensy}), shown in Fig.~\ref{fig:electronic_components}(b), which can communicate with a computer through a USB2 serial port. The microcontroller offers numerous digital pins for SPI communication as well as $2$ Digital-to-Analog (DAC) converters for generating analog signals. The board offers up to $4$ slots for RJ45 connectors which can be used to send SPI commands to the illumination modules through ethernet cables. Additionally, it offers up to $6$ slots for externally triggering capture devices through digital pulses, whose peak voltage is regulated by appropriate resistors. The Teensy $3.6$ supports a limited amount of storage memory on which a program capable of understanding the commands of the provided configuration file is pre-loaded. At the same time, it provides an accurate internal timer for sending signals at millisecond intervals.

\subsection{Software}
\label{subsec:software}
The software design aligns with the principles of \emph{flexibility} and \emph{modularity} described above. We have adopted a microservice architecture which uses REST APIs such that a process can send HTTP requests for capturing data from each available capture device. \vspace{0.1cm}

\paragraph*{\textbf{Device Servers}} Each capture device must follow a device server interface and should just implement a class providing methods for its \emph{initialization}, \emph{setting device parameters} and \emph{capturing a data sample}. This framework simplifies the process of adding new capture devices which only need to implement the aforementioned methods and are agnostic to the remaining system design. At the same time, for camera sensors (which are the ones used in our realization of the framework), it additionally provides a general camera capture device interface for reducing any additional software implementation needs. 
\vspace{0.1cm}

\paragraph*{\textbf{Configuration File}} The whole system's operation is determined by a JSON configuration file. It defines which capture devices and illumination sources will be used as well as the timestamps they will receive signals for their activation or deactivation. Further, it specifies initialization or runtime parameters for each capture device allowing adjustments to their operational characteristics without any software changes. As such, it can be used to fully determine a synchronized capture sequence between all available illumination sources and capture devices. Optionally, it can define a different preview sequence used for presenting data to the user through the GUI. Finally, it also determines the dataset names that will be used in the output HDF5 file to store the data from different capture devices. \vspace{0.1cm}

\paragraph*{\textbf{Graphical User Interface}} The GUI provides data preview and capture capabilities. In preview mode, it enters in a continuous loop of signals to all available capture devices and illumination sources and repeatedly sends HTTP requests to all underlying device servers while data is being previewed on the computer screen. In capture mode, it first sends a capture request to each capture device for a predefined number of frames dictated by the JSON configuration file and then puts the controller into a capture loop for sending the appropriate signals. Captured data is packaged into an HDF5 file and sent to a database for storage.

\subsection{Biometric Sensor Suites}
\label{sec:biometric_sensor_suites}

\begin{figure*}[!t]
    \centering
    \includegraphics{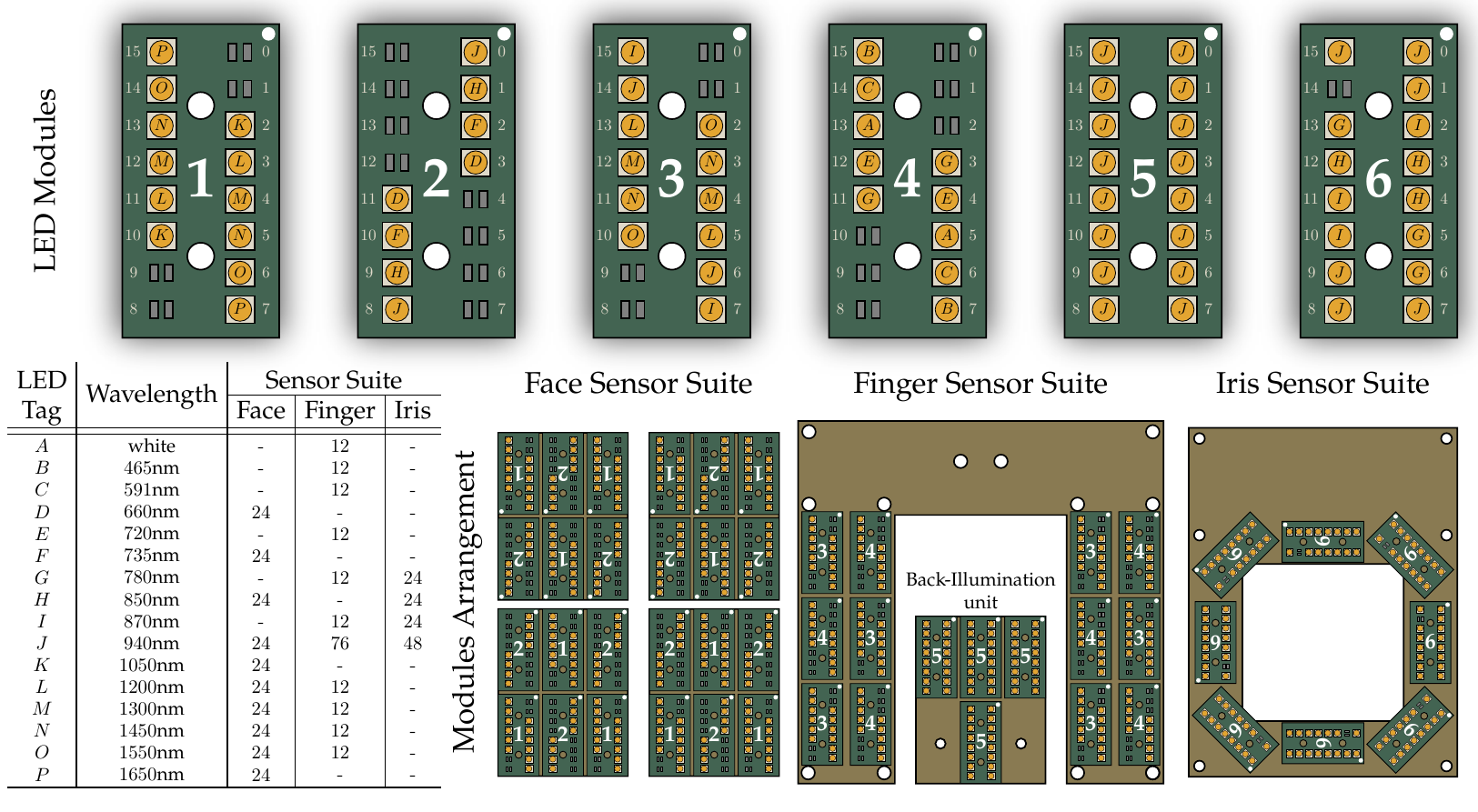}
    \caption{Main LED types and illumination modules used in the proposed biometric sensor suites. For each modality (\emph{face}, \emph{finger} or \emph{iris}), illumination modules are combined in different arrangements for achieving illumination uniformity on the observed biometric samples. Each group of modules can receive commands from the main controller board of Fig.~\ref{fig:system} through ethernet cables. Here, we refer to any separate LED tag ($A-P$) as representing a wavelength even though some of them might consist of multiple wavelengths (e.g., white light).}
    \label{fig:illumination_modules}
\end{figure*}

In this section, we dive into more details on the specifics of the realization of the presented framework on the \emph{face}, \emph{finger} and \emph{iris} biometric modalities. For our presented systems, all capture devices are cameras and all output data is frame sequences appropriately synchronized with the activation of particular light sources.

We use a variety of cameras each one sensitive to different portions (VIS, NIR, SWIR and LWIR or Thermal) of the electromagnetic spectrum. Table~\ref{tab:list_of_cameras} summarizes all cameras used in our system along with their main specifications. It is apparent that cameras share different characteristics in terms of their resolution, frame rate or dynamic range (bit depth). For some cameras, the sensitivity is restricted by using external band-pass filters in front of their lenses. The cameras were selected, among many options in the market, with the goal of balancing performance, data quality, user-friendliness and cost (but clearly different sensors could be selected based on the application needs). All cameras supporting hardware triggering operate in blocking-mode, i.e., waiting for trigger signals from the controller for a frame to be captured. This way, synchronized frames can be obtained. A few cameras (see Table~\ref{tab:list_of_cameras}) do not support hardware triggering and are synchronized using software countdown timers during the capture process. Even though this triggering mechanism is not millisecond accurate, the timestamps of each frame are also stored so that one can determine the closest frames in time to frames originating from the hardware triggered cameras.

For the illumination modules, we chose a variety of LEDs emitting light at different wavelengths covering a wide range of the spectrum. Here, without loss of generality, we will refer to any separate LED type as representing a wavelength even though some of them might consist of multiple wavelengths (e.g., white light). The choice of LEDs was based on previous studies on multispectral biometric data (as discussed in section~\ref{sec:introduction}) as well as cost and market availability of SMD LEDs from vendors (e.g., ~\cite{led_vendor1, led_vendor2, led_vendor3, led_vendor4}). For each biometric sensor suite, we tried to maximize the available wavelengths considering each LED's specifications and the system as a whole. Illumination modules are mounted in different arrangements on simple illumination boards containing an RJ45 connector for SPI communication with the main controller board through an ethernet cable. To achieve light uniformity, we created $6$ main types of illumination modules which attempt to preserve LED symmetry. Wavelength selection and module arrangement for each sensor suite is presented in Fig.~\ref{fig:illumination_modules}. In summary:

\begin{itemize}
    \item \textbf{\underline{Face sensor suite}}: Employs $10$ wavelengths mounted on $2$ types of illumination modules and arranged in $4$ separate groups. $24$ illumination modules with $240$ LEDs are used in total.
    \item \textbf{\underline{Finger sensor suite}}: Employs $11$ wavelengths mounted on $3$ types of illumination modules and arranged in $2$ separate groups. $16$ illumination modules with $184$ LEDs are used in total.
    \item \textbf{\underline{Iris sensor suite}}: Employs $4$ wavelengths mounted on a single illumination module type and arranged circularly. $8$ illumination modules with $120$ LEDs are used in total .
\end{itemize}

All system components are mounted using mechanical  parts~\cite{thorlabs} or custom-made 3D printed parts and enclosed in metallic casings~\cite{protocase, protocase_designer} for protection and user-interaction. Additionally, all lenses used (see Table~\ref{tab:list_of_cameras}) have a fixed focal length and each system has an optimal operating distance range based on the Field-of-View (FOV) and Depth-of-Field (DoF) of each camera-lens configuration used. It is important to note that our systems are prototypes and every effort was made to maximize efficiency and variety of captured data. However, the systems could be miniaturized using smaller cameras, fewer or alternate illumination sources or additional components, such as mirrors, for more compact arrangement and total form factor reduction. Such modifications would not interfere with the concepts of the proposed framework which would essentially remain the same.

\begin{figure*}[!htb]
    \centering
    \ifarxiv
    \includegraphics{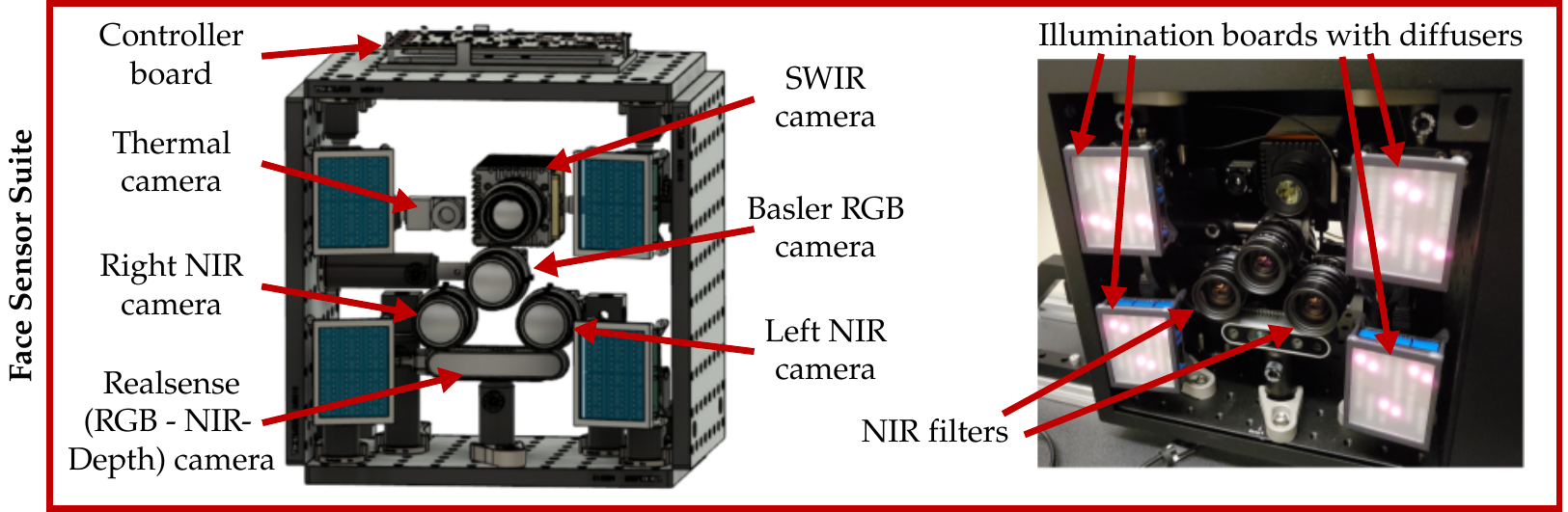}
    \else
    \includegraphics{Figures/FaceBiometricSuite_TBIOM.pdf}
    \fi
    \caption{Overview of the face sensor suite. Left side: 3D modeling of the system; Right side: Actual developed system.}
    \label{fig:face_biometric_suite}
\end{figure*}

\begin{figure*}[!htb]
    \centering
    \includegraphics{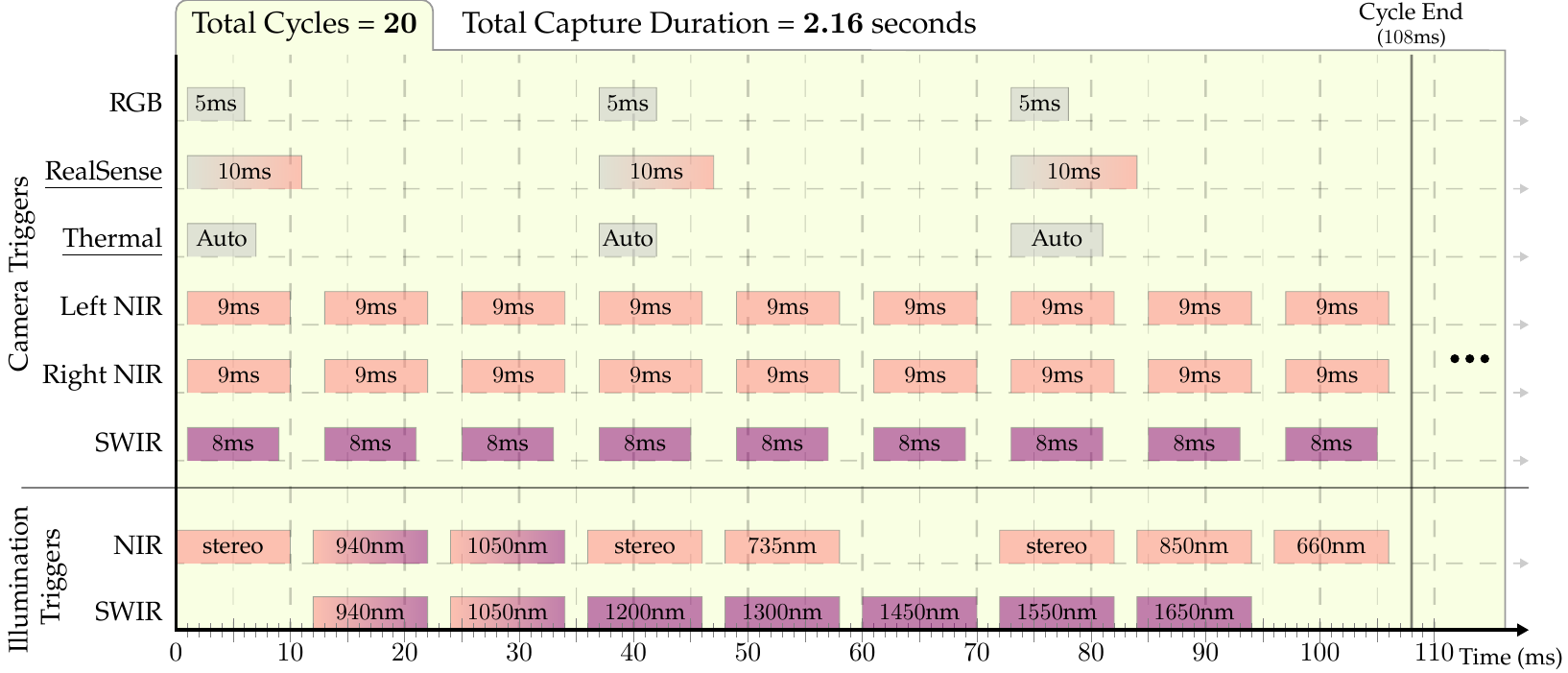}
    \caption{Face sensor suite synchronization sequence between cameras (software triggered cameras are underlined) and illumination sources. The width of each box represents the exposure time of each camera (or marked as ``Auto'' if auto-exposure is used) as well as the duration that each illumination source is active. The RGB, NIR and Depth channels of the RealSense~\cite{realsense} camera are internally synchronized to be captured at the same time. We capture $20$ cycles of the presented sequence for a total capture duration of $2.16$ seconds. The gray color represents cameras that are not affected by LED illumination, while the other two colors represent the sensitivity of each camera to the LED illumination sources. Finally, the NIR illumination denoted by ``stereo" refers to data captured at a constant frame rate and could be used for stereo depth reconstruction. In this configuration, ``stereo" data was captured using the $735$nm wavelength but multiple wavelengths could be simultaneously activated.}
    \label{fig:face_synchronization}
\end{figure*}

\subsection{Face Sensor Suite}
\label{subsec:face_sensor_suite}
\begin{figure*}[!t]
    \centering
    \ifarxiv
    \includegraphics{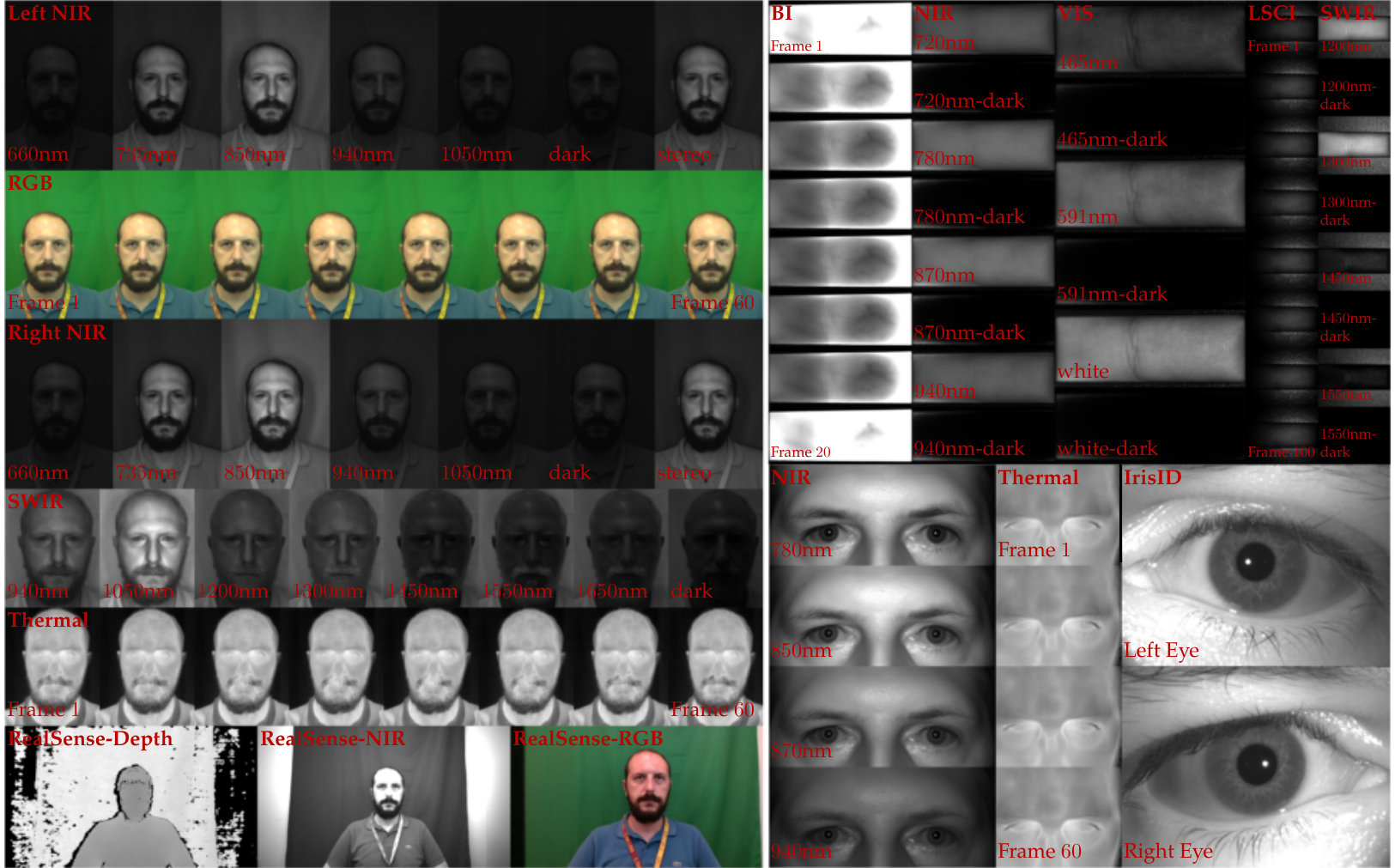}
    \else
    \includegraphics{Figures/Example_Captures_TBIOM.pdf}
    \fi
    \caption{Overview of captured data by the proposed sensor suites for \emph{face} (left), \emph{finger} (top-right) and \emph{iris} (bottom-right) biometric modalities. For cameras affected by LED illumination or capturing different data types, the middle frame of the capture sequence is shown. For the remaining cameras, equally spaced frames of the whole captured sequence are presented. Images are resized for visually pleasing arrangement and the relative size of images is not preserved.}
    \label{fig:example_captures}
\end{figure*}

The face sensor suite uses $6$ cameras capturing RGB, NIR~($\times 2$), SWIR, Thermal and Depth data as summarized in Table~\ref{tab:list_of_cameras}. An overview of the system is depicted in Fig.~\ref{fig:face_biometric_suite}. Except for the LED modules, we further use two big bright white lights on both sides of our system (not shown in the figure) to enable uniform lighting conditions for the RGB cameras. The subject sits in front of the system and the distance to the cameras is monitored by the depth indication of the RealSense camera~\cite{realsense}. We use a distance of $\sim 62$cm from the RealSense camera, which allows for good focus and best FOV coverage from most cameras. For the cameras affected by the LED illumination, we also capture frames when all LEDs are turned off, which can be used as ambient illumination reference frames. The synchronization sequence provided to the system through the JSON configuration file is presented in Fig.~\ref{fig:face_synchronization}. Finally, an overview of the captured data for a bona-fide sample is presented at the left side of Fig.~\ref{fig:example_captures} while an analysis of frames and storage needs is summarized in Table~\ref{tab:face_suite_data}. In this configuration, the system is capable of capturing $\sim 1.3$ GB of compressed data in $2.16$ seconds. Legacy compatible data is provided using either RGB camera of the system~\cite{basler_rgb, realsense}.

\begin{table}[!t]
\caption{Analysis of frames and storage needs for the data captured by the face sensor suite for a single subject. For the frames, we use notation (\emph{Number of frames} $\times$ \emph{Number of datasets in HDF5 file}). Each dataset corresponds to different illumination conditions for each data type.}
\label{tab:face_suite_data}
\centering
\begin{tabular}{r||cccc} 
\multirow{2}{*}{\textbf{Camera}} & \textbf{Data} &\textbf{Lit} & \textbf{Non-Lit} & \textbf{Bit} \\
 & \textbf{Type} & \textbf{Frames} & \textbf{Frames} & \textbf{Depth} \\
\toprule \hline
Basler RGB~\cite{basler_rgb} & RGB & $60 \times 1$ & $-$ & $8\times3$ \\
RealSense~\cite{realsense} & RGB &$60 \times 1$ & $-$ &$8 \times 3$\\
RealSense~\cite{realsense} & Depth & $60 \times 1$ & $-$ & $16$\\
RealSense~\cite{realsense} & NIR & $60 \times 1$ & $-$ & $8$ \\
Boson~\cite{boson_face} & Thermal & $-$ & $60 \times 1$ & $16$\\
Basler Left~\cite{basler_nir_face} & NIR &$20\times 6$ & $20 \times 1$ & $10$ (as $16$)\\
Basler Right~\cite{basler_nir_face} & NIR & $20\times 6$ & $20 \times 1$ & $10$ (as $16$)\\
Bobcat~\cite{bobcat} & SWIR &$20\times 7$ & $40 \times 1$ & $16$\\
\hline
\bottomrule
\multicolumn{5}{l}{\textbf{Storage}: $\sim 1.3$ GB, \quad \textbf{Capture Time}: $2.16$ seconds.}
\end{tabular}
\end{table}

\begin{figure*}[!htb]
    \centering
    \ifarxiv
    \includegraphics{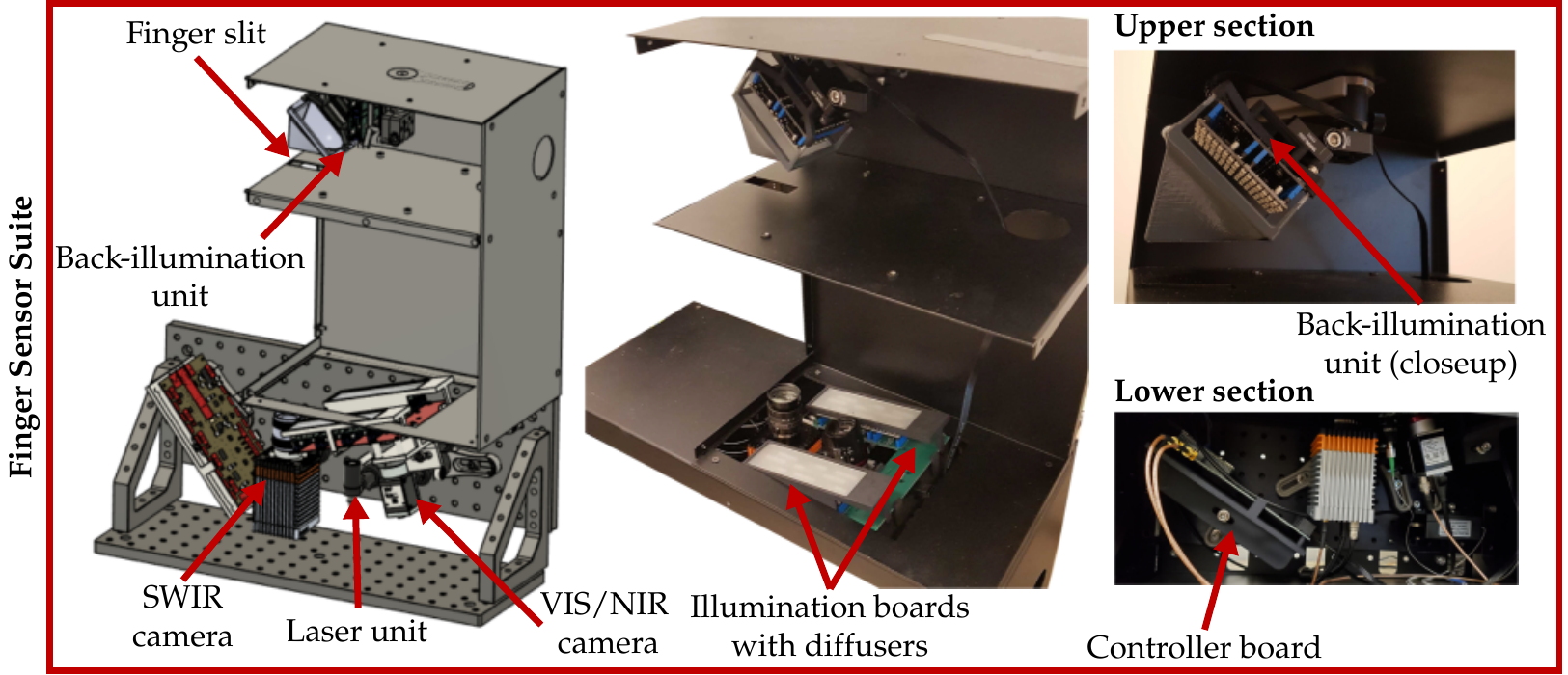}
    \else
    \includegraphics{Figures/FingerBiometricSuite_TBIOM.pdf}
    \fi
    \caption{Overview of the finger sensor suite. Left side: 3D modeling of the system; Remaining: Actual developed system.}
    \label{fig:finger_biometric_suite}
\end{figure*}

\begin{figure*}[!htb]
    \centering
    \includegraphics{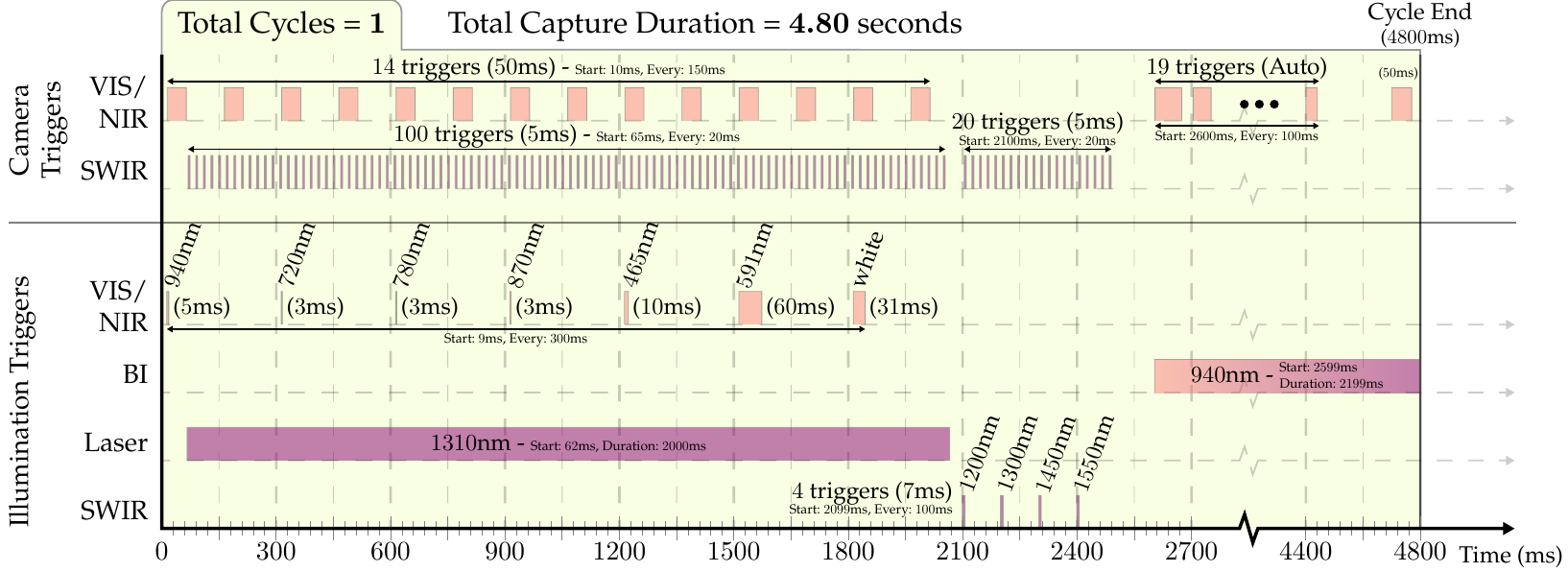}
    \caption{Finger sensor suite synchronization sequence between cameras and illumination sources. The width of each box represents the exposure time of each camera (or marked as ``Auto'' if auto-exposure is used) as well as the duration that each illumination source is active. We capture a single cycle of the presented sequence for a total capture duration of $4.80$ seconds. The colors represent the sensitivity of each camera to the illumination sources allowing simultaneous capture from both cameras in some instances.}
    \label{fig:finger_synchronization}
\end{figure*}

Looking closer at the face sensor suite, the $2$ NIR cameras constitute a stereo pair and can be used for high resolution 3D reconstruction of the biometric sample. Such an approach is not analyzed in this work. However, it requires careful calibration of the underlying cameras for estimating their intrinsic and extrinsic parameters. Moreover, despite face detection being a rather solved problem for RGB data~\cite{Bulat2018, yang2020fan}, this is not the case for data in different spectra. To enable face detection in all captured frames, we use a standard calibration process using checkerboards~\cite{Zhang2000}. For the checkerboard to be visible in all wavelength regimes, a manual approach is used when a sequence of frames is captured offline while the checkerboard is being lit with a bright halogen light. This makes the checkerboard pattern visible and detectable by all cameras which allows the standard calibration estimation process to be followed. The face can then be easily detected in the RGB space~\cite{Bulat2018, yang2020fan} and the calculated transformation for each camera can be applied to detect the face in the remaining camera frames.

\begin{table}[!htb]
\caption{Analysis of frames and storage needs for the data captured by the finger sensor suite for a single finger. For the frames, we use notation (\emph{Number of frames} $\times$ \emph{Number of datasets in HDF5 file}). Each dataset corresponds to different illumination conditions for each data type.}
\label{tab:finger_suite_data}
\centering
\begin{tabular}{r||cccc} 
\multirow{2}{*}{\textbf{Camera}} & \textbf{Data} &\textbf{Lit} & \textbf{Non-Lit} & \textbf{Bit} \\
 & \textbf{Type} & \textbf{Frames} & \textbf{Frames} & \textbf{Depth} \\
\toprule \hline
Basler~\cite{basler_nir_finger} & VIS/NIR  & $\;\;\:\, 1 \times 7$ & $1 \times 7$ & $12$ (as $16$)\\
Basler~\cite{basler_nir_finger} & BI       & $\;\, 20 \times 1$ & $-$ & $12$ (as $16$)\\
Bobcat~\cite{bobcat} & LSCI  & $100 \times 1$ & $-$ & $16$\\
Bobcat~\cite{bobcat} & SWIR  & $\;\;\;\, 1 \times 4$ & $4 \times 4$ & $16$\\
\hline \bottomrule
\multicolumn{5}{l}{\textbf{Storage}: $\sim 33$ MB, \quad \textbf{Capture Time}: $4.8$ seconds.}
\end{tabular}
\vspace{-0.2cm}
\end{table}

\subsection{Finger Sensor Suite}
\label{subsec:finger_sensor_suite}
The finger sensor suite uses $2$ cameras sensitive in the VIS/NIR and SWIR parts of the spectrum, as summarized in Table~\ref{tab:list_of_cameras}. An overview of the system is depicted in Fig.~\ref{fig:finger_biometric_suite}. The subject places a finger on the finger slit of size $15 \times 45$mm$^2$, facing downwards, which is imaged by the $2$ available cameras from a distance of $\sim 35$ cm. The finger sensor suite uses two additional distinct types of data compared to the remaining sensor suites, namely, Back-Illumination (BI) and Laser Speckle Contrast Imaging (LSCI). \vspace{0.1cm}

\paragraph*{\textbf{Back-Illumination}} Looking at Fig.~\ref{fig:finger_biometric_suite} and Fig.~\ref{fig:illumination_modules}, one can observe that the illumination modules are separated in two groups. The first one lies on the side of the cameras lighting the front side of the finger (front-illumination) while the second shines light atop the finger slit which we refer to as BI. This allows capturing images of the light propagating through the finger and can be useful for PAD by either observing light blockage by non-transparent materials used in common PAIs or revealing the presence of veins in a finger of a bona-fide sample. The selected NIR wavelength of $940$nm enhances penetration though the skin as well as absorption of light by the hemoglobin in the blood vessels~\cite{Raghavendra2014, Gupta2014, Wang2007, Kolberg2020} making them appear dark. Due to the varying thickness of fingers among different subjects, for BI images we use auto-exposure and capture multiple frames so intensity can be adjusted such that the captured image is not over-saturated nor under-exposed. \vspace{0.1cm}

\paragraph*{\textbf{Laser Speckle Contrast Imaging}} Apart from the incoherent LED illumination sources, the finger sensor suite also uses a coherent illumination source, specifically a laser at $1310$nm~\cite{laser}, which sends a beam at the forward part of the system's finger slit. The laser is powered directly by the power of the Teensy $3.6$~\cite{teensy} and its intensity can be controlled through an analog voltage using the DAC output of the controller board (as shown in Fig.~\ref{fig:system}). Illuminating a rough surface through a coherent illumination source leads to an interference pattern, known as speckle pattern. For static objects, the speckle pattern does not change over time. However, when there is motion (such as motion of blood cells through finger veins), the pattern changes at a rate dictated by the velocity of the moving particles and imaging this effect can be used for LSCI~\cite{Briers2013, Hussein2018, Keilbach2018, Mirzaalian2019, Sun2019}. The selected wavelength of $1310$nm enables penetration of light through the skin and the speckle pattern is altered over time as a result of the underlying blood flow for bona-fide samples. This time-related phenomenon can prove useful as an indicator of liveness and, in order to observe it, we capture a sequence of frames while the laser is turned on.

The synchronization sequence provided to the system through the JSON configuration file is presented in Fig.~\ref{fig:finger_synchronization}, where it is shown that complementary spectrum sensitivity of the utilized cameras is exploited for synchronous capture while enabling multiple illumination sources (e.g., laser and NIR light). For each type of data captured under the same lighting conditions and the same camera parameters (i.e., exposure time), we also capture frames when all LEDs are turned off which serve as ambient illumination reference frames. Finally, an overview of the captured data for a bona-fide sample is presented at the top-right part of Fig.~\ref{fig:example_captures} while an analysis of frames and storage needs per finger is summarized in Table~\ref{tab:finger_suite_data}. In this configuration, the system is capable of capturing $\sim 33$ MB of compressed data in $4.80$ seconds. Legacy compatible data is provided through the captured visible light images as we will show in section~\ref{subsec:legacy_compatibility}.

\begin{figure*}[!t]
    \centering
    \ifarxiv
    \includegraphics{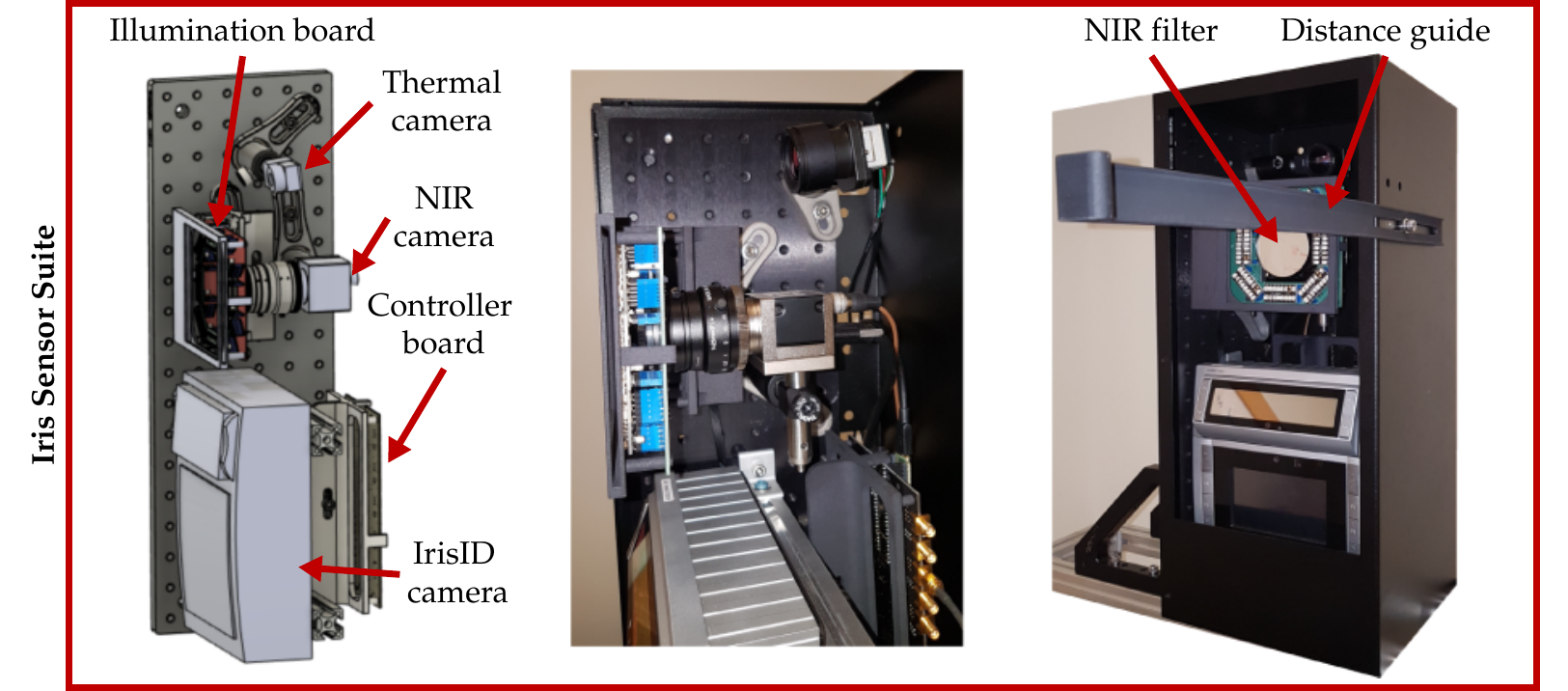}
    \else
    \includegraphics{Figures/IrisBiometricSuite_TBIOM.pdf}
    \fi
    \caption{Overview of the iris sensor suite. Left side: 3D modeling of the system; Remaining: Actual developed system.}
    \label{fig:iris_biometric_suite}
\end{figure*}

\begin{figure*}
    \centering
    \includegraphics{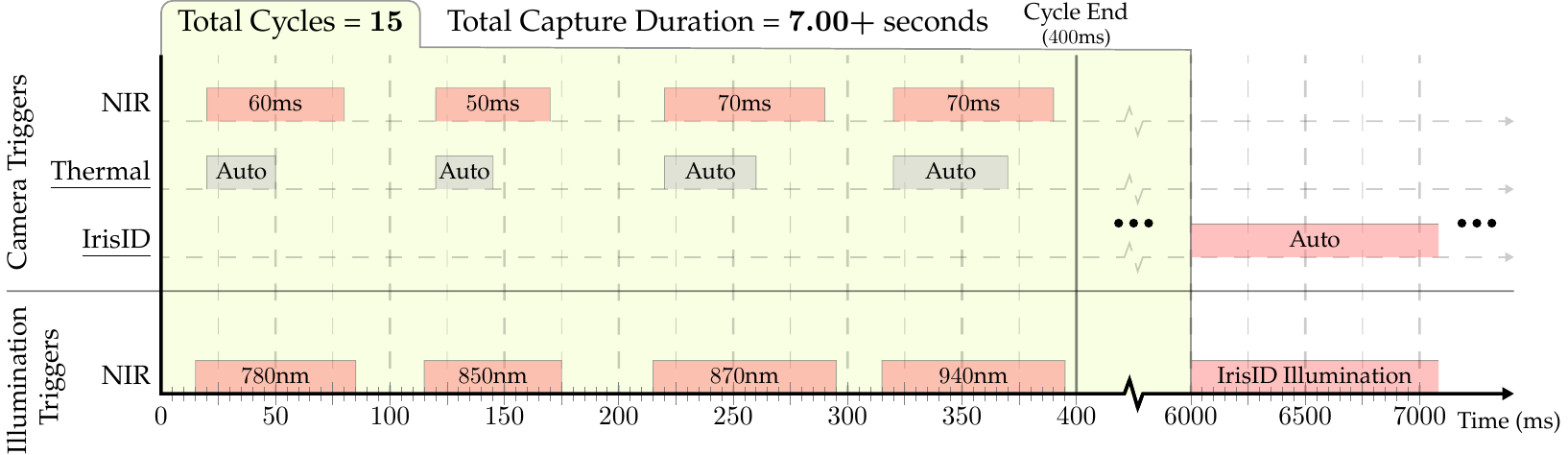}
    \caption{Iris sensor suite synchronization sequence between cameras (software triggered cameras are underlined) and illumination sources. The width of each box represents the exposure time of each camera (or marked as ``Auto'' if auto-exposure is used) as well as the duration that each illumination source is active. We capture $15$ cycles of the presented sequence and then enable the IrisID camera. The total capture duration is $7.00$ or more seconds (depending on the capture time of the IrisID camera which requires subject cooperation). The gray color represents cameras that are not affected by LED illumination, while the other color represents the sensitivity of each camera to the LED illumination sources.}
    \label{fig:iris_synchronization}
\end{figure*}

\subsection{Iris Sensor Suite}
\label{subsec:iris_sensor_suite}
The iris sensor suite uses $3$ cameras capturing NIR and Thermal data, as summarized in Table~\ref{tab:list_of_cameras}. An overview of the system is depicted in Fig.~\ref{fig:iris_biometric_suite}. The subject stands in front of the system at a distance of $\sim 35$ cm guided by the 3D printed distance guide on the right side of the metallic enclosure. The synchronization sequence provided to the system through the JSON configuration file is presented in Fig.~\ref{fig:iris_synchronization}. The IrisID camera~\cite{iris_id} employs its own NIR LED illumination and has an automated way of capturing data giving feedback and requiring user interaction. Hence, it is only activated at the end of the capture from the remaining $2$ cameras. An overview of the captured data for a bona-fide sample is presented at the bottom-right part of Fig.~\ref{fig:example_captures} while an analysis of frames and storage needs is summarized in Table~\ref{tab:iris_suite_data}. Note, that the IrisID provides the detected eyes directly while the remaining data require the application of an eye detection algorithm. For detecting eyes in the thermal images, we use the same calibration approach discussed in section~\ref{subsec:face_sensor_suite} where eyes can first be detected in the NIR domain and then their coordinates transformed to find the corresponding area in the thermal image. Both the data from the IrisID and the NIR camera are legacy compatible as we will show in section~\ref{subsec:legacy_compatibility}. Besides, the IrisID camera is one of the sensors most frequently used in the market. 

\begin{table}[!t]
\caption{Analysis of frames and storage needs for the data captured by the iris sensor suite for a single subject. For the frames, we use notation (\emph{Number of frames} $\times$ \emph{Number of datasets in HDF5 file}). Each dataset corresponds to different illumination conditions for each data type.}
\label{tab:iris_suite_data}
\centering
\begin{tabular}{r||cccc} 
\multirow{2}{*}{\textbf{Camera}} & \textbf{Data} &\textbf{Lit} & \textbf{Non-Lit} & \textbf{Bit} \\
 & \textbf{Type} &\textbf{Frames} & \textbf{Frames} & \textbf{Depth}\\
\toprule \hline
Basler~\cite{basler_nir_iris} & NIR & $15 \times 4$ & $-$ & $12$ (as $16$)\\
Boson~\cite{boson_iris} & Thermal & $-$ &  $60 \times 1$ & $16$\\
\multirow{2}{*}{IrisID~\cite{iris_id}} & \multirow{2}{*}{NIR} & $2 \times 1$ & \multirow{2}{*}{$-$} & \multirow{2}{*}{$8$} \\
& &  ($1$ per eye) & & \\
\hline \bottomrule
\multicolumn{5}{l}{\textbf{Storage}: $\sim 0.46$ GB, \quad \textbf{Capture Time}: $6$ seconds plus IrisID} \\
\multicolumn{5}{l}{capture time (\textbf{max}: 25 seconds).}
\end{tabular}
\end{table}

One of the drawbacks of the current iris sensor suite is its sensitivity to the subject's motion and distance due to the rather narrow DoF of the utilized cameras/lenses as well as the long exposure time needed for acquiring bright images. As a result, it requires careful operator feedback to the subject for appropriate positioning in front of the system. Higher intensity illumination or narrow angle LEDs could be used to combat this problem by further closing the aperture of the cameras so that the DoF is increased. However, further research is required for this purpose, taking into consideration possible eye-safety concerns, not present in the current design which employs very low energy LEDs.
\section{Experiments}
\label{sec:experiments}
In our analysis so far, we have verified the principles of \emph{flexibility} and \emph{modularity} governing the system design in our proposed framework. In this section, we focus on the principles of \emph{legacy compatibility} and \emph{complementarity} of the captured data and showcase that they can provide rich information when applied to PAD. The main focus of our work is to present the flexible multispectral biometrics framework and not devise the best performing algorithm for PAD since the captured data can be used in a variety of ways for obtaining the best possible performance. Instead, we follow a data-centric approach and attempt to understand the contribution of distinct regimes of the multispectral data towards detecting different types of PAIs.

\subsection{Datasets}
\label{subsec:dataset}
We have held $7$ data collections with the proposed systems. However, our systems have undergone multiple improvements throughout this period and some data is not fully compatible with the current version of our system (see for example the previous version of our finger sensor suite in~\cite{Hussein2018}, which has been largely simplified here). A series of publications have already used data from earlier versions of our systems (see~\cite{Nikisins2019, Kotwal2019, Jaiswal2019, George2020} for \emph{face} and~\cite{Hussein2018, Keilbach2018, Tolosana2018, Mirzaalian2019, Barrero2019, Sun2019, Tolosana2020, Kolberg2020} for \emph{finger}).

The datasets used in our analysis contain only data across data collections that are compatible with the current design (i.e., the same cameras, lenses and illumination sources, as the ones described in section~\ref{sec:system_design}, were used). They involve $5$ separate data collections of varying size, demographics and PAI distributions that were performed using $2$ distinct replicas of our systems in $5$ separate locations (leading to possibly different ambient illumination conditions and slight modifications in the positions of each system's components). Participants presented their biometric samples at least twice to our sensors and a few participants engaged in more than one data collections. Parts of the data will become publicly available through separate publications and the remaining could be distributed later by the National Institute of Standards and Technology (NIST)~\cite{nist}.

In this work, we separate all data from the aforementioned data collections in two groups (data from the $4$ former data collections and data from the last data collection). The main statistics for the two groups which will be referred to as \emph{Dataset I} and \emph{Dataset II}, respectively, as well as their union (\emph{Combined}) are summarized in Table~\ref{tab:datasets}. The reason for this separation is twofold. First, we want to study a cross-dataset analysis scenario for drawing general conclusions. Second, during the last data collection, data was also captured using a variety of existing commercial sensors for \emph{face}, \emph{finger} and \emph{iris}. Therefore, \emph{Dataset II} constitutes an ideal candidate on which the \emph{legacy compatibility} principle of our proposed sensor suites can be analyzed.

\begin{table}[!t]
    \centering
    \caption{Studied datasets and their union. For each biometric modality, we group PAIs into broader categories (marked in gray) and present the number of samples and PAI species (sp.) included in each. Not all available PAI species are included in this categorization. PAI categories whose appearance depends heavily on the subject and preparation method are marked with $^*$. Finally, the contact lens (CL) category marked with $^{\dagger}$ groups contact lenses whose specific type is unknown or their count in the dataset is small for being separately grouped.}
    \label{tab:datasets}
    \includegraphics[]{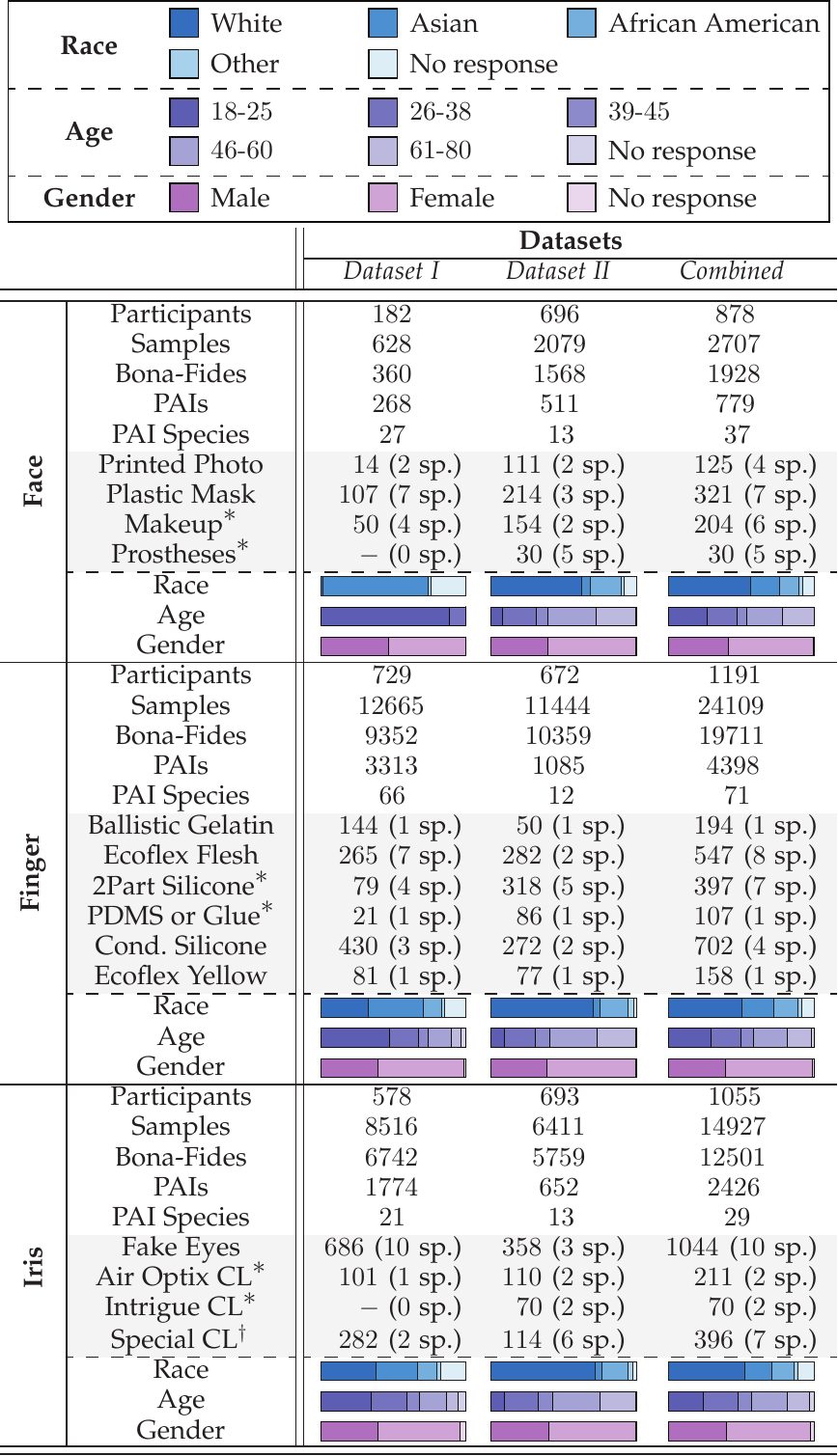}
\end{table}

\begin{figure*}[!htb]
    \centering
    \ifarxiv
    \includegraphics[width=\linewidth]{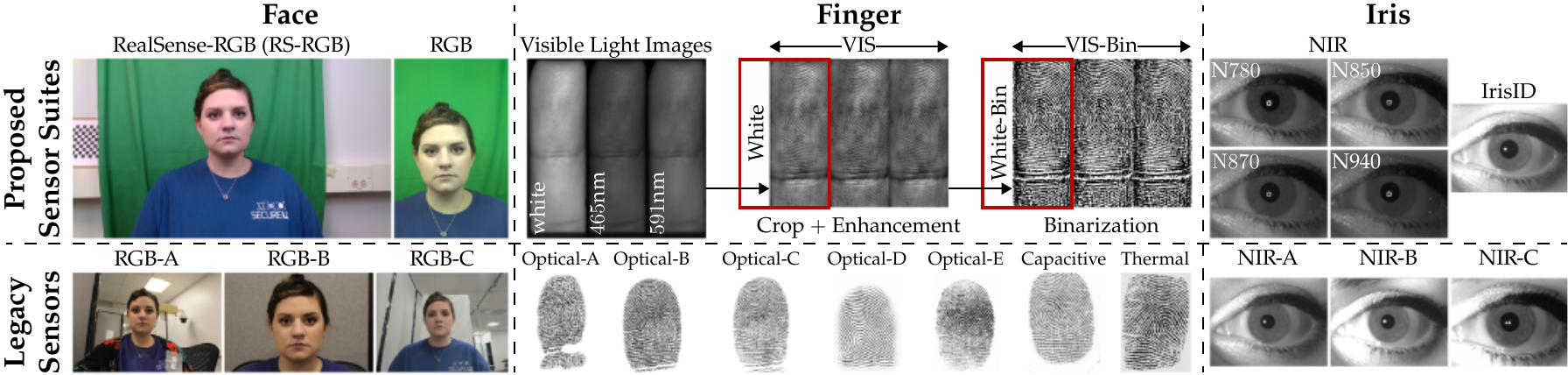}
    \else
    \includegraphics[width=\linewidth]{Figures/Legacy_Images_Examples.pdf}
    \fi
    \caption{Examples of legacy compatible data captured by the proposed sensor suites and multiple legacy sensors, retrieved from \emph{Dataset II} (see Table~\ref{tab:datasets}). All data correspond to the same participant while \emph{finger} and \emph{iris} images depict the right index finger and left eye, respectively. For each data type, the figure further presents the notation used in Tables~\ref{tab:enrollment_rates} and~\ref{tab:match_rates}.}
    \label{fig:legacy_samples}
\end{figure*}

For each biometric modality, we define a set of PAI categories (see Table~\ref{tab:datasets}), which will be helpful for our analysis. As observed, multiple PAI species are omitted from the categorization. We tried to form compact categories, which encapsulate different PAI characteristics, as well as consider cases of unknown PAI categories among the two datasets. Finally, it is important to note that the age and race distributions of the participants among the two datasets is drastically different. \emph{Dataset I} is dominated by young people of Asian origin while \emph{Dataset II} includes a larger population of Caucasians or African Americans with a skewed age distribution toward older ages, especially for \emph{face} data.

%
%

\begin{table}[!t]
    \centering
    \caption{Bona-fide enrollment rates for each sensor used in \emph{Dataset II} (see Fig.~\ref{fig:legacy_samples} and Table~\ref{tab:datasets}), calculated using Neurotechnology's SDK software~\cite{neurotechnology_sdk} for a minimum quality threshold of $40$. The third column lists the enrollment rate when all samples are considered while the fourth presents the corresponding enrollment rate when enrollment of at least one sample per participant and \emph{BP} is considered a success. Similarly, the last two columns list the total bona-fide samples and unique participant-\emph{BP} bona-fide samples per sensor, respectively. The best enrollment rates per biometric modality are highlighted in bold.}
    \label{tab:enrollment_rates}
    \resizebox{\columnwidth}{!}{
    \begin{tabular}{c||ccccc} 
        & \multirow{2}{*}{\textbf{Sensor}} & 
        \textbf{Enroll.} &
        \textbf{Unique} &
        \textbf{Total} &
        \textbf{Unique} \\
        & & \textbf{Rate} & \textbf{Enroll. Rate} & \textbf{Samples} & \textbf{Samples} \\ \toprule \hline
            
        \multirow{5}{*}{\rotatebox[origin=c]{90}{\textbf{Face}}} & RGB-A & $95.51\%$ & $95.76 \%$ & $690$ & $661$ \\
        & RGB-B & $\mathbf{98.91}\%$ & $98.66\%$ & $642$ & $522$ \\
        & RGB-C & $95.81\%$ & $95.21\%$ & $645$ & $522$ \\
        & \cellcolor{Gray}{RGB} & \cellcolor{Gray}{$98.60\%$} & \cellcolor{Gray}{$\mathbf{99.42}\%$} & \cellcolor{Gray}{$1568$} & \cellcolor{Gray}{$521$} \\ 
        & \cellcolor{Gray}{RS-RGB} & \cellcolor{Gray}{$98.47\%$} & \cellcolor{Gray}{$99.23\%$} & \cellcolor{Gray}{$1568$} & \cellcolor{Gray}{$521$} \\ \hline
            
        \multirow{11}{*}{\rotatebox[origin=c]{90}{\textbf{Finger}}} & Optical-A & $91.70\%$ & $93.08\%$ & $13416$ & $6405$ \\
        & Optical-B & $87.98\%$ & $89.26\%$ & $13316$ & $6399$ \\
        & Optical-C & $89.80\%$ & $88.41\%$ & $10440$ & $6350$ \\ 
        & Optical-D & $\mathbf{98.49}\%$ & $\mathbf{99.28}\%$ & $13805$ & $6423$ \\
        & Optical-E & $84.48\%$ & $84.06\%$ & $8757$ & $5922$ \\
        & Capacitive & $88.24\%$ & $91.63\%$ & $12905$ & $6378$ \\
        & Thermal & $86.59\%$ & $88.81\%$ & $11838$ & $6372$ \\
        & \cellcolor{Gray}{White} & \cellcolor{Gray}{$75.87\%$} & \cellcolor{Gray}{$79.64\%$} & \cellcolor{Gray}{$10359$} & \cellcolor{Gray}{$4932$} \\
        & \cellcolor{Gray}{VIS} & \cellcolor{Gray}{$78.70\%$} & \cellcolor{Gray}{$82.14\%$} & \cellcolor{Gray}{$10359$} & \cellcolor{Gray}{$4932$} \\
        & \cellcolor{Gray}{White-Bin}  & \cellcolor{Gray}{$80.75\%$} & \cellcolor{Gray}{$83.17\%$} & \cellcolor{Gray}{$10359$} & \cellcolor{Gray}{$4932$} \\
        & \cellcolor{Gray}{VIS-Bin} & \cellcolor{Gray}{$82.76\%$} & \cellcolor{Gray}{$85.00\%$} & \cellcolor{Gray}{$10359$} & \cellcolor{Gray}{$4932$} \\ \hline 
            
        \multirow{8}{*}{\rotatebox[origin=c]{90}{\textbf{Iris}}} & NIR-A & $93.12\%$ & $96.42\%$ & $1454$ & $1314$ \\
        & NIR-B & $96.58\%$ & $97.53\%$ & $1723$ & $1295$ \\
        & NIR-C & $98.06\%$ & $98.09\%$ & $1549$ & $1258$ \\ 
        & \cellcolor{Gray}{IrisID} & \cellcolor{Gray}{$\mathbf{98.54}\%$} & \cellcolor{Gray}{$\mathbf{99.23}\%$} & \cellcolor{Gray}{$5759$} & \cellcolor{Gray}{$1293$} \\
        & \cellcolor{Gray}{N$780$} & \cellcolor{Gray}{$94.44\%$} & \cellcolor{Gray}{$96.83\%$} & \cellcolor{Gray}{$5759$} & \cellcolor{Gray}{$1293$} \\
        & \cellcolor{Gray}{N$850$} & \cellcolor{Gray}{$98.14\%$} & \cellcolor{Gray}{$99.15\%$} & \cellcolor{Gray}{$5759$} & \cellcolor{Gray}{$1293$} \\
        & \cellcolor{Gray}{N$870$} & \cellcolor{Gray}{$93.96\%$} & \cellcolor{Gray}{$97.22\%$} & \cellcolor{Gray}{$5759$} & \cellcolor{Gray}{$1293$} \\
        & \cellcolor{Gray}{N$940$} & \cellcolor{Gray}{$49.66\%$} & \cellcolor{Gray}{$60.94\%$} & \cellcolor{Gray}{$5759$} & \cellcolor{Gray}{$1293$} \\ \hline \bottomrule
    \end{tabular}}
\end{table}

%
%

\begin{table}[!t]
    \centering
    \caption{Bona-fide match rates between legacy compatible data provided by the proposed sensor suites and each one of the available legacy sensors in \emph{Dataset II} (see Table~\ref{tab:datasets}). Table entries correspond to the FNMR at $0.01\%$ FMR for each sensor pair, calculated using~\cite{neurotechnology_sdk}, with the highest match rates highlighted in bold. Only bona-fide samples for each participant and \emph{BP} that were enrolled by both sensors in each sensor pair were considered. For comparison, the average match rates between the data from \emph{finger} legacy sensors are: {Optical-A ({$1.91\%$), Optical-B ($2.49\%$), Optical-C ($\mathbf{1.78}\%$), Optical-D ($3.30\%$), Optical-E ($2.62\%$), Capacitive ($2.76\%$), Thermal ($3.32\%$).}}}
    \label{tab:match_rates}
    \begin{tabular}{c|c||cccccccccccccccccccc}
    \multicolumn{1}{c}{} & \textbf{Legacy} & \multicolumn{20}{c}{\textbf{Proposed}} \\ 
    \multicolumn{1}{c}{} & \textbf{Sensors} & \multicolumn{20}{c}{\textbf{Sensor Suites}} \\ \toprule \hline
    \multicolumn{1}{c}{} & & \multicolumn{10}{c}{RGB} & \multicolumn{10}{c}{RS-RGB} \\ \hline \hline
    \multirow{3}{*}{\rotatebox[origin=c]{90}{\textbf{Face}}} & RGB-A & 
    \multicolumn{10}{c}{$\mathbf{0.00} \%$} & \multicolumn{10}{c}{$0.07 \%$} \\
    & RGB-B & \multicolumn{10}{c}{$\mathbf{0.00} \%$} & \multicolumn{10}{c}{$0.01 \%$}  \\
    & RGB-C & \multicolumn{10}{c}{$\mathbf{0.00} \%$} & \multicolumn{10}{c}{$0.09 \%$} \\ \hline \hline
    
    \multicolumn{1}{c}{} & & \multicolumn{5}{c}{White} & \multicolumn{5}{c}{White-Bin} & \multicolumn{5}{c}{VIS} & \multicolumn{5}{c}{VIS-Bin} \\ \hline \hline
    \multirow{7}{*}{\rotatebox[origin=c]{90}{\textbf{Finger}}} & Optical-A & \multicolumn{5}{c}{$2.31 \%$} & \multicolumn{5}{c}{$\mathbf{2.19} \%$} & \multicolumn{5}{c}{$2.47 \%$} & \multicolumn{5}{c}{$2.47 \%$} \\
     & Optical-B & \multicolumn{5}{c}{$2.75 \%$} & \multicolumn{5}{c}{$3.09 \%$} & \multicolumn{5}{c}{$\mathbf{2.73} \%$} & \multicolumn{5}{c}{$3.28 \%$} \\
    & Optical-C & \multicolumn{5}{c}{$2.18 \%$} & \multicolumn{5}{c}{$2.38 \%$} & \multicolumn{5}{c}{$\mathbf{2.06} \%$} & \multicolumn{5}{c}{$2.32 \%$} \\
    & Optical-D & \multicolumn{5}{c}{$5.45 \%$} & \multicolumn{5}{c}{$\mathbf{5.00} \%$} & \multicolumn{5}{c}{$5.55 \%$} & \multicolumn{5}{c}{$5.17 \%$} \\
    & Optical-E & \multicolumn{5}{c}{$3.18 \%$} & \multicolumn{5}{c}{$\mathbf{3.07} \%$} & \multicolumn{5}{c}{$3.19 \%$} & \multicolumn{5}{c}{$3.31 \%$} \\
    & Capacitive & \multicolumn{5}{c}{$\mathbf{4.21} \%$} & \multicolumn{5}{c}{$4.58 \%$} & \multicolumn{5}{c}{$4.41 \%$} & \multicolumn{5}{c}{$4.84 \%$} \\
    & Thermal & \multicolumn{5}{c}{$\mathbf{6.41} \%$} & \multicolumn{5}{c}{$6.95 \%$} & \multicolumn{5}{c}{$6.71 \%$} & \multicolumn{5}{c}{$7.25 \%$} \\ \hline \hline
    
    \multicolumn{1}{c}{} & & \multicolumn{4}{c}{IrisID} & \multicolumn{4}{c}{N$780$} & \multicolumn{4}{c}{N$850$} & \multicolumn{4}{c}{N$870$} & \multicolumn{4}{c}{N$940$} \\ \hline \hline
    \multirow{3}{*}{\rotatebox[origin=c]{90}{\textbf{Iris}}} & NIR-A & \multicolumn{4}{c}{$\mathbf{0.67} \%$} & \multicolumn{4}{c}{$0.89\%$ } & \multicolumn{4}{c}{$0.89 \%$} & \multicolumn{4}{c}{$0.72 \%$} & \multicolumn{4}{c}{$1.92 \%$} \\
    & NIR-B & \multicolumn{4}{c}{$0.83 \%$} & \multicolumn{4}{c}{$0.94 \%$} & \multicolumn{4}{c}{$0.93 \%$} & \multicolumn{4}{c}{$0.65 \%$} & \multicolumn{4}{c}{$\mathbf{0.59} \%$} \\
    & NIR-C & \multicolumn{4}{c}{$1.18 \%$} & \multicolumn{4}{c}{$1.19 \%$} & \multicolumn{4}{c}{$1.12 \%$} & \multicolumn{4}{c}{$\mathbf{0.86} \%$} & \multicolumn{4}{c}{$1.21 \%$} \\ \hline \bottomrule
    
    \end{tabular}
\end{table}

\begin{figure*}[!htb]
    \centering
    \includegraphics[width=\linewidth]{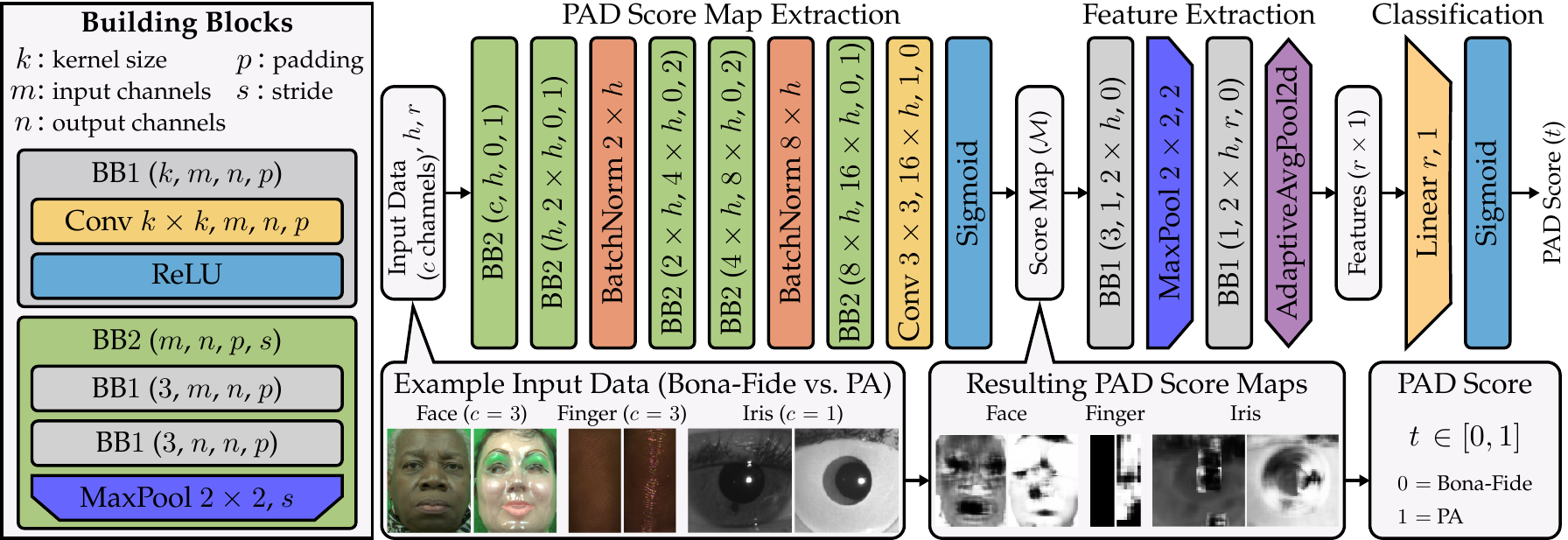}
    \caption{FCN Model architecture (extension of~\cite{Spinoulas2020}): Given parameter $h$ and number of features $r$, an input image of $c$ channels is first converted into a two dimensional PAD score map ($\cal{M}$) whose spatial distribution is then used to extract $r$ features and deduce the final PAD score $t \in [0,1]$ through a linear layer. Actual score map examples for bona-fide and PA samples are presented at the bottom part of the illustration, following the flow of the presented architecture.}
    \label{fig:model_architecture}
\end{figure*}

\subsection{Legacy Compatibility}
\label{subsec:legacy_compatibility}
As discussed above, during collecting \emph{Dataset II}, data from each participant was also collected using a variety of legacy sensors ($3$ different sensor types for \emph{face} and \emph{iris} and $7$ for \emph{finger}). Sponsor approval is required to release specific references for the legacy sensors used. Instead we provide descriptive identifiers, based on the data types each sensor captures. We now perform a comprehensive list of experiments to understand the legacy compatibility capabilities of our systems. For this purpose, we employ Neurotechnology's SDK software~\cite{neurotechnology_sdk}, which is capable of performing biometric data enrollment and matching. We use notation \emph{BP} (\emph{Biometric Position}) to refer to a specific sample (i.e., face, left or right eye or specific finger of the left or right hand of a subject). From our sensor suites, legacy compatible data for \emph{face} and \emph{iris} is used as is. For \emph{finger}, we noticed that the software was failing to enroll multiple high-quality samples, possibly due to the non-conventional nature of the captured finger images and, as a result, we considered two pre-processing steps. First, we cropped a fixed area of the captured image containing mostly the top finger knuckle and enhanced the image using adaptive histogram equalization. Second, we binarize the enhanced image using edge preserving noise reduction filtering and local adaptive thresholding. Fig.~\ref{fig:legacy_samples} provides an overview of data samples from all sensors, along with the notation used for each, and depicts the pre-processing steps for \emph{finger} data.

Using the SDK, we first perform enrollment rate analysis for all bona-fide samples in \emph{Dataset II} using $40$ as the minimum acceptable quality threshold. Following, we consider each pair between the proposed sensors and available legacy sensors and perform biometric template matching among all bona-fide samples for all participants with at least one sample for the same \emph{BP} enrolled from both sensors. The enrollment rate results are provided in Table~\ref{tab:enrollment_rates} and the match rates for each sensor pair are extracted by drawing a Receiver Operatic Characteristic (ROC) curve and reporting the value of False Non-Match Rate (FNMR) at $0.01\%$ False Match Rate (FMR)~\cite{VijayaKumar2011} in Table~\ref{tab:match_rates}. For \emph{finger}, we analyze the performance of white light images as well as the performance when all $3$ visible light images are used (see Fig.~\ref{fig:legacy_samples}) and any one of them is enrolled. When matching, the image with the highest enrollment quality is used.

\begin{figure*}
    \centering
    \ifarxiv
    \includegraphics[width=\linewidth]{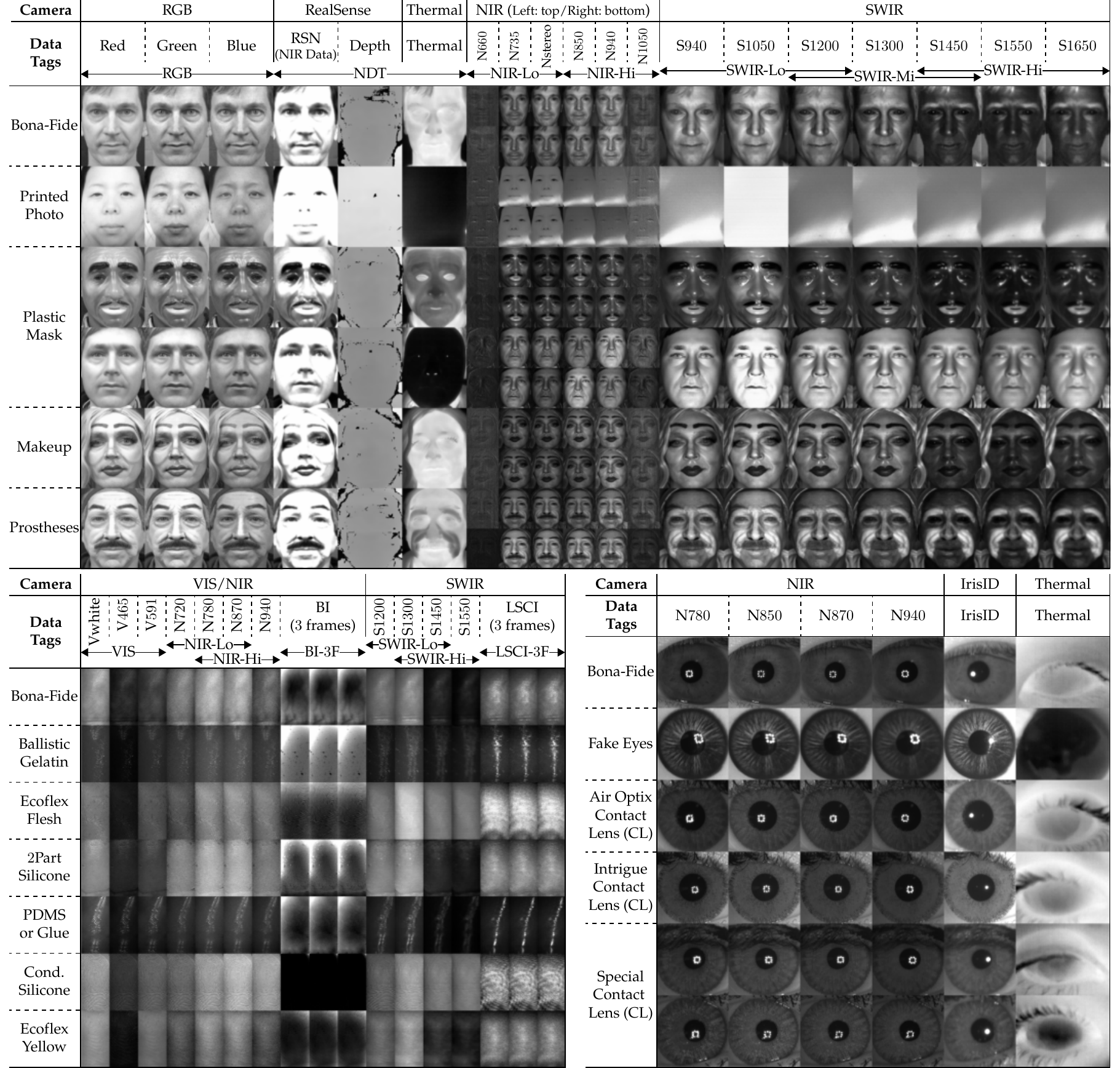}
    \else
    \includegraphics[width=\linewidth]{Figures/Data_Preprocessing_Examples.pdf}
    \fi
    \caption{Examples of pre-processed multispectral data for bona-fide samples and the main PAI categories defined in Table~\ref{tab:datasets} for each biometric modality. In some cases, images have been min-max normalized within each spectral regime for better visualization. The notation used in the figure is crucial for understanding the results in Fig.~\ref{fig:results_plots} and Table~\ref{tab:results_table}.}
    \label{fig:data_preprocessing}
\end{figure*}

From the results in the tables, it is apparent that the \emph{face} and \emph{iris} sensor suites provide at least one image type that is fully legacy compatible. For the \emph{finger} data, enrollment rate appears to be sub-optimal while match rates are in some cases on par with the average match rates between legacy sensors (compare with values in caption of Table~\ref{tab:match_rates}). However, the utilized analysis software proves very sensitive to the input image type and the same images when binarized (compare White vs. White-Bin and VIS vs. VIS-Bin entries in Table~\ref{tab:enrollment_rates}) exhibit $\sim 5\%$ increase in enrollment rates. Hence, we are confident that a more careful selection of pre-processing steps~\cite{Hara2009} or the use of an alternate matching software could lead to improved performance. Besides, the Optical-D legacy sensor, despite covering the smallest finger area and having the lowest resolution among all analyzed legacy sensors, seems to outperform the others by a large margin, indicating the high sensitivity of the enrollment and matching software to selected parameters. Deeper investigation into this topic, however, falls out of the scope of this work.

\subsection{Presentation Attack Detection}
\label{subsec:presentation_attack_detection}
In order to support the \emph{complementarity} principle of our design, we devise a set of PAD experiments for each biometric modality. Two class classification, with labels $\{0,1\}$ assigned to bona-fide and PA samples, respectively, is performed using a convolutional neural network (CNN) based model. \vspace{0.1cm}

\paragraph*{\textbf{Model Architecture}}
Due to the limited amounts of training data, inherent in biometrics, we follow a patch-based approach where each patch in the input image is first classified with a PAD score in $[0, 1]$ and then individual scores are fused to deduce the final PAD score $t \in [0,1]$ for each sample. Unlike traditional patch-based approaches where data is first extracted for patches of a given size and stride and then passed through the network (e.g., \cite{Hussein2018, Mirzaalian2019}), we use an extension of the fully-convolutional-network (FCN) architecture presented in~\cite{Spinoulas2020}, as depicted in Fig.~\ref{fig:model_architecture}.
The network consists of $3$ parts:

\begin{enumerate}
    \item \underline{Score Map Extraction}: Assigns a value in $[0, 1]$ to each patch producing a score map ($\cal{M})$ through a set of convolutions and non-linearities while batch normalization layers are used to combat over-fitting.
    \item \underline{Feature Extraction}: Extracts $r$ score map features through a shallow CNN. 
    \item \underline{Classification}: Predicts the final PAD score $t$ by passing the score map features through a linear layer.
\end{enumerate}

The suggested network architecture was inspired by the supervision channel approach in~\cite{Jourabloo2018, Liu2018} and its first part (identical to~\cite{Spinoulas2020}) is equivalent to a patch-based architecture when the stride is $1$, albeit with increased computational efficiency and reduced memory overhead. A drawback of the FCN architecture compared to a genuine patch-based model, however, is that patches of a sample image are processed together and the batch size needs to be smaller, reducing intra-variability in training batches. The two remaining parts, instead of just performing score averaging, consider the spatial distribution of the score map values for deducing the final PAD score, as shown in the examples at the bottom part of Fig.~\ref{fig:model_architecture} for a bona-fide and PA sample per modality. The additional feature extraction and classification layers were considered due to the possible non-uniformity of PAIs especially in the case of \emph{face} and \emph{iris} data, unlike \emph{finger} data~\cite{Spinoulas2020}, where a PAI usually covers the whole finger image area passed to the network. \vspace{0.1cm}

\paragraph*{\textbf{Training Loss}} The network architecture in Fig.~\ref{fig:model_architecture} guarantees ${\cal{M}}_i \in [0,1],\; i = 0, \dots, N_{\cal{M}}-1$, where $N_{\cal{M}}$ is the total number of elements in $\cal{M}$, through the sigmoid layer. However, it does not guarantee that $\cal{M}$ would represent an actual PAD score map for the underlying sample. In order to enforce that all patches within each sample belong to the same class, we employ pixel-wise supervision on $\cal{M}$ such that ${\cal{M}}_i = g,\; i = 0, \dots, N_{\cal{M}}-1$ where $g \in \{0,1\}$ is the ground truth label of the current sample. Denoting the Binary Cross-Entropy loss function as ${\cal{B}}(x, y)$ the sample loss $\cal{L}$ is calculated as: 

\begin{equation}
\label{eq:training_loss}
    {\cal{L}} ={\cal{B}}(t,g) + \frac{w}{N_{{\cal{M}}}}\sum_{i=0}^{N_{{\cal{M}}}-1}{\cal{B}}({\cal{M}}_i,g),
\end{equation}
where $w \ge 0$ is a constant weight.

\subsection{Presentation Attack Detection Experiments}

As discussed earlier, the goal of our work is the understanding of the contribution of each spectral channel or regime to the PAD problem as well as the strengths and weaknesses of each type of data by following a data-centric approach.  Therefore, we use a model that remains the same across all compared experiments per modality. As such, we try to gain an understanding on how performance is solely affected by the data rather than the number of trainable model parameters, specific model architecture or other training hyperparameters. We first summarize the data pre-processing and training protocols used in our experiments and then describe the experiments in detail. \vspace{0.1cm}

\paragraph*{\textbf{Data Pre-processing}}
The data for each biometric modality is pre-processed as follows, where any resizing operation is performed using bicubic interpolation:

\begin{itemize}
    \item \underline{\emph{Face}}: Face landmarks are detected in the RGB space using~\cite{yang2020fan} and the bounding box formed by the extremities is expanded by $25\%$ toward the top direction. The transformations obtained by the calibration process described in section~\ref{subsec:face_sensor_suite} are then used to warp each image channel to the RGB image dimensions and the bounding box area is cropped. Finally, all channels are resized to $320 \times 256$ pixels. A single frame from the captured sequence is used per sample.
    \item \underline{\emph{Finger}}: A fixed region of interest is cropped per channel such that the covered finger area is roughly the same among all cameras (based on their resolution, system geometry and dimensions of the finger slit mentioned in section~\ref{subsec:finger_sensor_suite}). The cropped area covers the top finger knuckle which falls on an almost constant position for all samples, since each participant uses the finger slit for presenting each finger. Finally, all channels are resized to $160 \times 80$ pixels.
    \item \underline{\emph{Iris}}: For the data captured using the NIR and IrisID cameras, Neurotechnology's SDK~\cite{neurotechnology_sdk} is employed for performing iris segmentation. The iris bounds are then used as a region of interest for cropping. Each image is finally resized to $256 \times 256$ pixels. For the thermal data, we use the whole eye region (including the periocular area). The center of the eye is extracted from the segmentation calculated on the NIR camera's images and the corresponding area in the Thermal image is found by applying the calculated transformation between the two cameras (as discussed in section~\ref{subsec:iris_sensor_suite}). The cropped area is finally resized to $120 \times 160$ pixels. A single multispectral frame from the captured sequence is used per sample. We always use the frame with the highest quality score provided by~\cite{neurotechnology_sdk} during segmentation. If segmentation fails for all available frames, the sample is discarded.
\end{itemize}

Exploiting the camera synchronization in our systems, for \emph{face} and \emph{iris} data which rely on geometric transformations, the particular frames extracted from each channel are the closest ones in time to the reference frame where face or eye detection was applied (based on each frame's timestamps). For all biometric modalities, if dark channel frames are available for any spectral channel (see Fig.~\ref{fig:example_captures}), the corresponding time-averaged dark channel is first subtracted. The data is then normalized in $[0, 1]$ using the corresponding channel's bit depth (see Tables~\ref{tab:face_suite_data},~\ref{tab:finger_suite_data},~\ref{tab:iris_suite_data}). Examples of pre-processed data for bona-fide samples and the PAI categories defined in Table~\ref{tab:datasets} are presented in Fig.~\ref{fig:data_preprocessing}. In some cases, images have been min-max normalized within each spectral regime for better visualization. The notations used in the figure will become important in the following analysis.
\vspace{0.1cm}

\begin{table}[!t]
    \centering
    \caption{Parameters used for all experiments.}
    \label{tab:training_parameters}
    \begin{tabular}{cc}
        \toprule \hline
        \multicolumn{2}{c}{\textbf{Model/Loss Parameters}} \\ \hline
        $h = 16$, $r = 4\times h$, $w = 10$ & \multirow{2}{*}{(see Fig.~\ref{fig:model_architecture} and Eq.~(\ref{eq:training_loss}))} \\ 
        $c\,$: Defined by input image channels & \\ \hline \hline 
        \multicolumn{2}{c}{\textbf{Training Parameters}} \\ \hline
        \multicolumn{2}{c}{\emph{Initial learning rate}: $2 \times 10^{-4}$,
        \emph{Minimum learning rate}: $1 \times 10^{-7}$,} \\
        \multicolumn{2}{c}{\emph{Learning rate scheduling}: Reduce by $0.5$ on validation loss} \\
        \multicolumn{2}{c}{plateau with (patience: $10$ epochs, threshold: $1 \times 10^{-4}$),} \\
        \multicolumn{2}{c}{\emph{Optimizer}: Adam~\cite{Kingma2014}, \emph{Epochs}: $100$, \emph{Batch Size}: $16$} \\ \hline \bottomrule
    \end{tabular}
\end{table}

\paragraph*{\textbf{Training Protocols}} We follow two different training protocols using the datasets presented in Table~\ref{tab:datasets}:

\begin{enumerate}
    \item \underline{\emph{3Fold}}: All data from the \emph{Combined} dataset is divided in $3$ folds. For each fold, the training, validation and testing sets consist of $55\%$, $15\%$ and $30\%$ of data, respectively. The folds were created considering the participants such that no participant appears in more than one set leading to slightly different percentages than the aforementioned ones. 
    \item \underline{\emph{Cross-Dataset}}: \emph{Dataset I} is used for training and validation ($85\%$ and $15\%$ of data, respectively) while \emph{Dataset II} is used for testing. In this scenario, a few participants do appear in both datasets, for the \emph{finger} and \emph{iris} cases, but their data was collected at a different point in time, a different location and using a different replica of our biometric sensor suites (see participant counts in Table~\ref{tab:datasets}).
\end{enumerate}
\vspace{0.1cm}

\begin{figure*}[!t]
    \centering
    \includegraphics[width=\linewidth]{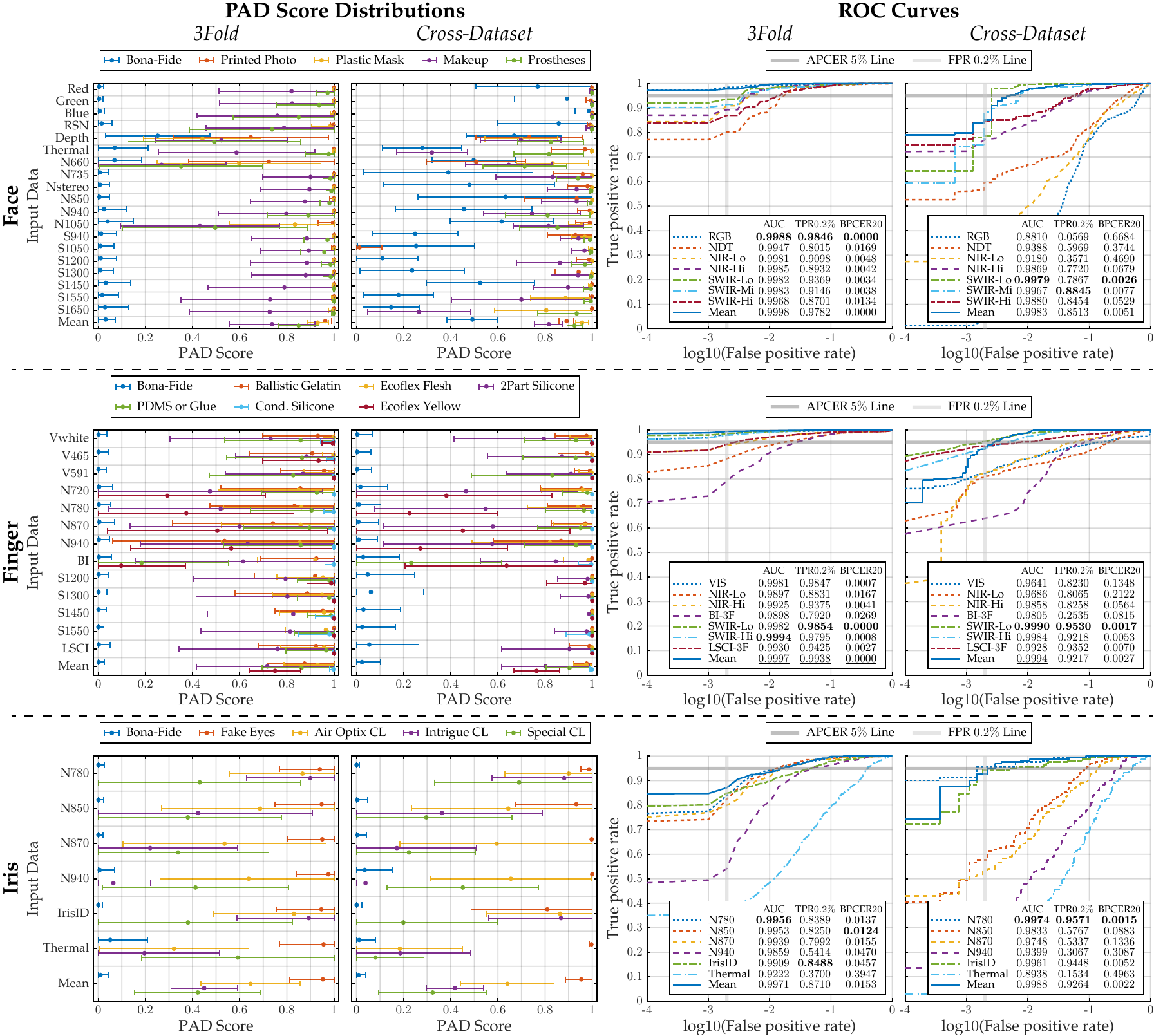}
    \caption{Left: PAD score distributions for single channel experiments; Right: ROC curves corresponding to $3$-channel experiments for \emph{face} and \emph{finger} and single channel experiments for \emph{iris}. In the ROC curve legends, the best performance is highlighted in bold while the score fusion result (Mean) is underlined when outperforming the best individual experiment.}
    \label{fig:results_plots}
\end{figure*}

We now conduct a series of comprehensive experiments to analyze the PAD performance capabilities of the captured data. First, for all three biometric modalities, we perform experiments when each spectral channel is used separately as input to the model of Fig.~\ref{fig:model_architecture} (i.e., $c=1$). For \emph{face} and \emph{finger} data, due to the large number of channels, we further conduct experiments when combinations of $c = 3$ input channels are used. On one hand, this approach aids in summarizing the results in a compact form but also constitutes a logical extension. For \emph{face}, $3$ is the number of channels provided by the RGB camera while for \emph{finger}, there are $3$ visible light illumination sources and LSCI data is inherently time-dependent, hence sequential frames are necessary for observing this effect. We choose not to study larger channel combinations so that we accentuate the individual contribution of each type of available data to the PAD problem, but always adhere to the rule of comparing experiments using the same number of input channels and therefore contain the same amount of trainable model parameters. 

Each experiment uses the same model and training parameters, summarized in Table~\ref{tab:training_parameters}. During training, each channel is standardized to zero-mean and unit standard deviation based on the statistics of all images in the training set, while the same normalizing transformation is applied when testing. All experiments are performed on both (\emph{3Fold} and \emph{Cross-Dataset}) training protocols explained above. The notation used for each individual channel and each triplet combination in the experiments is illustrated in Fig.~\ref{fig:data_preprocessing}. For each type of experiment, we also calculate the performance of the mean PAD score fusion of all individual experiments (denoted as Mean). As performance metrics, we report the Area Under the Curve (AUC), the True Positive Rate at $0.2\%$ False Positive Rate (denoted as TPR$0.2\%$) and the Bona-fide Presentation Classification Error Rate at a fixed Attack Presentation Classification Error Rate (APCER) of $5\%$ (denoted as BPCER$20$ in the ISO~\cite{iso} standard).

The results from all experiments are summarized in Fig.~\ref{fig:results_plots} and Table~\ref{tab:results_table}. The left part of Fig.~\ref{fig:results_plots} analyzes the single channel experiments by drawing error bars of the PAD score distributions for bona-fide samples and each PAI category defined in Table~\ref{tab:datasets}. The error bars depict the mean and standard deviation of each score distribution bounded by the PAD score limits $[0,1]$. Hence, full separation of error bars between bona-fides and PAIs does not imply perfect score separation. However, it can showcase in a clear way which channels are most effective at detecting specific PAI categories. The right part of Fig.~\ref{fig:results_plots} presents the calculated ROC curves and relevant performance metrics for the $3$-channel experiments for \emph{face} and \emph{finger} and $1$-channel experiments for \emph{iris}. The same results are then re-analyzed per PAI category in Table~\ref{tab:results_table} by calculating each performance metric for an ROC curve drawn by considering only bona-fide samples and a single PAI category each time. The table is color coded such that darker shades denote performance degradation and helps at interpreting the underlying metric values. The $[0,1]$ grayscale values were indeed selected using the average value between AUC, TPR$0.2\%$ and ($1-$ BPCER$20$) for each colored entry. It is important to note that for the \emph{iris} experiments, only samples for which iris segmentation was successful in all channels are used in the analysis for a fair comparison among the different models.

\begin{table*}[!htb]
    \centering
    \caption{Performance metric analysis per PAI category. For each experiment from the ROC curve's section in Fig.~\ref{fig:results_plots}, separate ROC curves are extracted considering only bona-fide samples and samples from a single PAI category (as defined in Table~\ref{tab:datasets}). The table is color coded such that darker shades denote reduction in performance. The $[0,1]$ grayscale values were selected using the average value between AUC, TPR$0.2\%$ and ($1-$ BPCER$20$) for each colored entry. The best performance per PAI category and training protocol is highlighted in bold while the score fusion result (Mean) is underlined when matching or outperforming the best individual experiment.}
    \label{tab:results_table}
    \includegraphics[width=\linewidth]{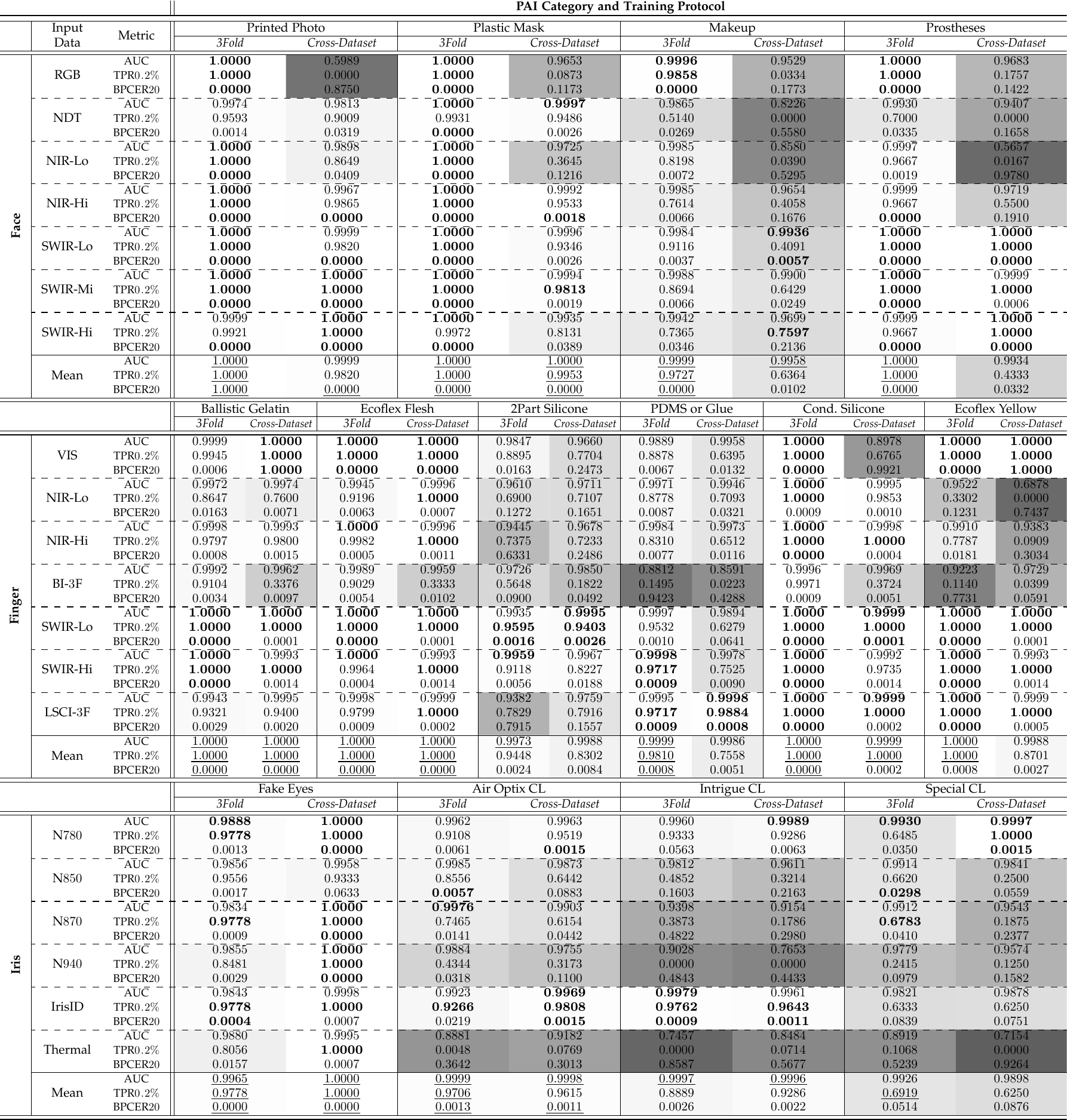}
\end{table*}

By analyzing the presented results, we make the following observations:
\begin{itemize}
    \item Some channels behave exactly as expected by the human visual perception of the images (e.g., Thermal channel success on Plastic Mask and Fake Eyes).
    \item Certain PAI categories appear to be easily detectable by most channels (e.g., Ballistic Gelatin and Ecoflex Flesh) while others (e.g., Prostheses and PDMS or Glue) exhibit consistent separation when SWIR/LSCI illumination is used, supporting the \emph{complementarity} principle of the proposed system.
    \item The \emph{complementarity} principle is further strengthened by the performance of simple score averaging (denoted as Mean), which is the best in multiple cases.
    \item The \emph{Cross-Dataset} protocol performance appears to be severely degraded for channels where cameras are affected by ambient illumination conditions (e.g., visible or NIR light). This is particularly apparent in the \emph{face} experiments where RGB data performance changes from best to worst between the two training protocols and a huge PAD score shift can be observed in the score distributions. This effect is also aggravated by the smaller size of the \emph{face} dataset which can lead to overfitting as well as the vastly different demographics between the training and testing sets (discussed in section~\ref{subsec:dataset}). On the contrary, higher wavelength channels appear to be more resilient to both ambient illumination and demographic variations, consistent with the existing literature~\cite{Steiner2016}.
    \item In a few cases, \emph{Cross-Dataset} protocol evaluation outperforms the equivalent \emph{3Fold} experiments. This can be explained due the smaller variety of PAI species in \emph{Dataset II} as well as the larger variety of bona-fide samples in the \emph{Combined} dataset, some of which might be inherently harder to classify correctly.
    \item For the \emph{iris} data, use of multispectral data seems to be less important. The Thermal channel, while successful at detecting fake eyes, appears weak at detecting PAI contact lenses (indeed multiple eyes of participants wearing regular contact lenses are misclassified due to the darker appearance of their pupil-iris area). At the same time, only a single NIR channel appears to have slightly better performance than the IrisID camera which also uses NIR illumination, possibly due to its higher image resolution (see Fig.~\ref{fig:data_preprocessing}). Nevertheless, the fact that Mean score performance is sometimes superior suggests that certain samples are classified correctly mainly due to the multispectral nature of the data. However, as discussed in section~\ref{subsec:iris_sensor_suite}, the current iris system setup is not optimal and suffers from motion blur as well as cross-wavelength blur when a participant is moving during capture. Blurriness can obscure small details necessary for the detection of contact lens PAIs. Indeed, the N$780$ channel, which exhibits the highest performance, was the one that was in best focus and whose images were usually receiving the highest quality scores during the legacy compatibility analysis of section~\ref{subsec:legacy_compatibility}.
\end{itemize}

In general, the analysis suggests that for each biometric modality, there is channels which can alone offer high PAD performance. Clearly, some of the problems observed in the \emph{Cross-Dataset} protocol analysis could be alleviated by using pre-training, transfer learning or fine-tuning techniques, but the purpose of our work is to emphasize on the limitations originating from certain wavelength regimes and stress the importance of the availability of a variety of spectral bands for training robust classification models. Besides, models using a larger input channel stack can further enhance PA detection, as shown in~\cite{George2020,Tolosana2020,Spinoulas2020,Hussein2018, Mirzaalian2019}.

\section{Conclusion}
\label{sec:conclusion}
In this work, we presented a multispectral biometrics system framework along with its realization on \emph{face}, \emph{finger} and \emph{iris} biometric modalities. We described the proposed systems in detail and explained how they adhere to the principles of \emph{flexibility}, \emph{modularity}, \emph{legacy compatibility} and \emph{complementarity}. Further, we showcased that the captured data can provide rich and diverse information useful at distinguishing a series of presentation attack instrument types from bona-fide samples. The variety of synchronized biometric data captured through the proposed systems can open doors to various different applications. We hope that our work will spur further research in this domain and make multispectral biometric data a commodity for researchers to experiment with. We believe that multispectral data for biometrics can be one of the key ingredients for detecting future and ever more sophisticated presentation attacks.
\ifCLASSOPTIONcompsoc
  \section*{Acknowledgments}
\else
  \section*{Acknowledgment}
\fi
\label{sec:acknowledgments}

This research is based upon work supported by the Office of the Director of National Intelligence (ODNI), Intelligence Advanced Research Projects Activity (IARPA), via IARPA R\&D Contract No. 2017-17020200005. The views and conclusions contained herein are those of the authors and should not be interpreted as necessarily representing the official policies or endorsements, either expressed or implied, of the ODNI, IARPA, or the U.S. Government. The U.S. Government is authorized to reproduce and distribute reprints for Governmental purposes notwithstanding any copyright annotation thereon.

The authors would like to thank Marta Gomez-Barrero, Jascha Kolberg and Christoph Busch for their helpful discussions and contributions on the finger biometric modality.
\ifCLASSOPTIONcaptionsoff
  \newpage
\fi



\bibliographystyle{IEEEtran}
\bibliography{references}

\begin{thebibliography}{10}
\providecommand{\url}[1]{#1}
\csname url@samestyle\endcsname
\providecommand{\newblock}{\relax}
\providecommand{\bibinfo}[2]{#2}
\providecommand{\BIBentrySTDinterwordspacing}{\spaceskip=0pt\relax}
\providecommand{\BIBentryALTinterwordstretchfactor}{4}
\providecommand{\BIBentryALTinterwordspacing}{\spaceskip=\fontdimen2\font plus
\BIBentryALTinterwordstretchfactor\fontdimen3\font minus
  \fontdimen4\font\relax}
\providecommand{\BIBforeignlanguage}[2]{{%
\expandafter\ifx\csname l@#1\endcsname\relax
\typeout{** WARNING: IEEEtran.bst: No hyphenation pattern has been}%
\typeout{** loaded for the language `#1'. Using the pattern for}%
\typeout{** the default language instead.}%
\else
\language=\csname l@#1\endcsname
\fi
#2}}
\providecommand{\BIBdecl}{\relax}
\BIBdecl

\bibitem{Spurny2015}
J.~{Spurný}, M.~{Doleel}, O.~{Kanich}, M.~{Drahanský}, and K.~{Shinoda},
  ``{New Materials for Spoofing Touch-Based Fingerprint Scanners},'' in
  \emph{2015 International Conference on Computer Application Technologies},
  2015, pp. 207--211.

\bibitem{Lafkih2015}
M.~{Lafkih}, P.~{Lacharme}, C.~{Rosenberger}, M.~{Mikram}, S.~{Ghouzali}, M.~E.
  {Haziti}, W.~{Abdul}, and D.~{Aboutajdine}, ``Application of new alteration
  attack on biometric authentication systems,'' in \emph{2015 First
  International Conference on Anti-Cybercrime (ICACC)}, 2015, pp. 1--5.

\bibitem{ba_under_threat}
``{Biometric Authentication Under Threat: Liveness detection Hacking},''
  \url{https://www.blackhat.com/us-19/briefings/schedule/}.

\bibitem{Marcel2019}
\BIBentryALTinterwordspacing
S.~Marcel, M.~S. Nixon, J.~Fi{\'{e}}rrez, and N.~W.~D. Evans, Eds.,
  \emph{Handbook of Biometric Anti-Spoofing - Presentation Attack Detection,
  Second Edition}, ser. Advances in Computer Vision and Pattern
  Recognition.\hskip 1em plus 0.5em minus 0.4em\relax Springer, 2019. [Online].
  Available: \url{https://doi.org/10.1007/978-3-319-92627-8}
\BIBentrySTDinterwordspacing

\bibitem{Munir2019}
\BIBentryALTinterwordspacing
R.~Munir and R.~A. Khan, ``{An extensive review on spectral imaging in
  biometric systems: Challenges $\&$ advancements},'' \emph{Journal of Visual
  Communication and Image Representation}, vol.~65, p. 102660, 2019. [Online].
  Available:
  \url{http://www.sciencedirect.com/science/article/pii/S1047320319302810}
\BIBentrySTDinterwordspacing

\bibitem{Roui-Abidi2009}
\BIBentryALTinterwordspacing
B.~Roui-Abidi and M.~Abidi, \emph{Multispectral and Hyperspectral
  Biometrics}.\hskip 1em plus 0.5em minus 0.4em\relax Boston, MA: Springer US,
  2009, pp. 993--998. [Online]. Available:
  \url{https://doi.org/10.1007/978-0-387-73003-5_163}
\BIBentrySTDinterwordspacing

\bibitem{Steiner2016}
\BIBentryALTinterwordspacing
H.~Steiner, S.~Sporrer, A.~Kolb, and N.~Jung, ``Design of an {Active}
  {Multispectral} {SWIR} {Camera} {System} for {Skin} {Detection} and {Face}
  {Verification},'' \emph{Journal of Sensors}, vol. 2016, p.~16, 2016.
  [Online]. Available: \url{http://dx.doi.org/10.1155/2016/9682453}
\BIBentrySTDinterwordspacing

\bibitem{Sigrononi2019}
A.~Signoroni, M.~Savardi, A.~Baronio, and S.~Benini, ``{D}eep {L}earning meets
  {H}yperspectral {I}mage {A}nalysis: {A} multidisciplinary review,''
  \emph{Journal of Imaging}, vol.~5, no.~5, 2019.

\bibitem{Engelsma2019}
J.~J. {Engelsma}, K.~{Cao}, and A.~K. {Jain}, ``{RaspiReader: Open Source
  Fingerprint Reader},'' \emph{IEEE Transactions on Pattern Analysis and
  Machine Intelligence}, vol.~41, no.~10, pp. 2511--2524, 2019.

\bibitem{Venkatesh2019}
S.~{Venkatesh}, R.~{Ramachandra}, K.~{Raja}, and C.~{Busch}, ``A new
  multi-spectral iris acquisition sensor for biometric verification and
  presentation attack detection,'' in \emph{2019 IEEE Winter Applications of
  Computer Vision Workshops (WACVW)}, 2019, pp. 47--54.

\bibitem{Zhang2016}
\BIBentryALTinterwordspacing
D.~Zhang, Z.~Guo, and Y.~Gong, \emph{Multispectral Biometrics Systems}.\hskip
  1em plus 0.5em minus 0.4em\relax Cham: Springer International Publishing,
  2016, pp. 23--35. [Online]. Available:
  \url{https://doi.org/10.1007/978-3-319-22485-5_2}
\BIBentrySTDinterwordspacing

\bibitem{realsense}
``{Intel\textsuperscript{\textregistered}
  RealSense\textsuperscript{\texttrademark} Depth Camera D435},''
  \url{https://www.intelrealsense.com/depth-camera-d435/}.

\bibitem{lumidigm}
``{HID\textsuperscript{\textregistered}
  Lumidigm\textsuperscript{\textregistered} V-Series Fingerprint Readers,
  v302-40},''
  \url{https://www.hidglobal.com/products/readers/single-finger-readers/lumidigm-v-series-fingerprint-readers}.

\bibitem{vistaey2}
``{Vista Imaging, Inc., VistaEY2 Dual Iris $\&$ Face Camera},''
  \url{https://www.vistaimaging.com/biometric_products.html}.

\bibitem{Chingovska2016}
\BIBentryALTinterwordspacing
I.~Chingovska, N.~Erdogmus, A.~Anjos, and S.~Marcel, \emph{Face Recognition
  Systems Under Spoofing Attacks}.\hskip 1em plus 0.5em minus 0.4em\relax Cham:
  Springer International Publishing, 2016, pp. 165--194. [Online]. Available:
  \url{https://doi.org/10.1007/978-3-319-28501-6_8}
\BIBentrySTDinterwordspacing

\bibitem{Raghavendra2017}
R.~{Raghavendra}, K.~B. {Raja}, S.~{Venkatesh}, F.~A. {Cheikh}, and C.~{Busch},
  ``On the vulnerability of extended multispectral face recognition systems
  towards presentation attacks,'' in \emph{2017 IEEE International Conference
  on Identity, Security and Behavior Analysis (ISBA)}, 2017, pp. 1--8.

\bibitem{Agarwal2017}
A.~{Agarwal}, D.~{Yadav}, N.~{Kohli}, R.~{Singh}, M.~{Vatsa}, and A.~{Noore},
  ``Face presentation attack with latex masks in multispectral videos,'' in
  \emph{2017 IEEE Conference on Computer Vision and Pattern Recognition
  Workshops (CVPRW)}, 2017, pp. 275--283.

\bibitem{Hussein2018}
M.~E. {Hussein}, L.~{Spinoulas}, F.~{Xiong}, and W.~{Abd-Almageed},
  ``Fingerprint presentation attack detection using a novel multi-spectral
  capture device and patch-based convolutional neural networks,'' in \emph{2018
  IEEE International Workshop on Information Forensics and Security (WIFS)},
  Dec 2018, pp. 1--8.

\bibitem{Agnieszka2018}
\BIBentryALTinterwordspacing
A.~Jenerowicz, P.~Walczykowski, L.~Gladysz, and M.~Gralewicz, ``{Application of
  hyperspectral imaging in hand biometrics},'' in \emph{Counterterrorism, Crime
  Fighting, Forensics, and Surveillance Technologies II}, H.~Bouma, R.~Prabhu,
  R.~J. Stokes, and Y.~Yitzhaky, Eds., vol. 10802, International Society for
  Optics and Photonics.\hskip 1em plus 0.5em minus 0.4em\relax SPIE, 2018, pp.
  129 -- 138. [Online]. Available: \url{https://doi.org/10.1117/12.2325489}
\BIBentrySTDinterwordspacing

\bibitem{Brauers2008}
J.~{Brauers}, N.~{Schulte}, and T.~{Aach}, ``Multispectral filter-wheel
  cameras: Geometric distortion model and compensation algorithms,'' \emph{IEEE
  Transactions on Image Processing}, vol.~17, no.~12, pp. 2368--2380, 2008.

\bibitem{Wu2020}
\BIBentryALTinterwordspacing
X.~Wu, D.~Gao, Q.~Chen, and J.~Chen, ``Multispectral imaging via nanostructured
  random broadband filtering,'' \emph{Opt. Express}, vol.~28, no.~4, pp.
  4859--4875, Feb 2020. [Online]. Available:
  \url{http://www.opticsexpress.org/abstract.cfm?URI=oe-28-4-4859}
\BIBentrySTDinterwordspacing

\bibitem{led_driver}
``{PCA9745B LED driver},''
  \url{https://www.digikey.com/product-detail/en/nxp-usa-inc/PCA9745BTWJ/568-14156-1-ND/9449780}.

\bibitem{teensy}
``{Teensy 3.6},'' \url{https://www.pjrc.com/store/teensy36.html}.

\bibitem{basler_rgb}
``{Basler acA1920-150uc},''
  \url{https://www.baslerweb.com/en/products/cameras/area-scan-cameras/ace/aca1920-150uc/}.

\bibitem{basler_nir_face}
``{Basler acA1920-150um},''
  \url{https://www.baslerweb.com/en/products/cameras/area-scan-cameras/ace/aca1920-150um/}.

\bibitem{basler_nir_iris}
``{Basler acA4096-30um},''
  \url{https://www.baslerweb.com/en/products/cameras/area-scan-cameras/ace/aca4096-30um/}.

\bibitem{basler_nir_finger}
``{Basler acA1300-60gmNIR},''
  \url{https://www.baslerweb.com/en/products/cameras/area-scan-cameras/ace/aca1300-60gmnir/}.

\bibitem{iris_id}
``{IrisID iCAM-7000 series, iCAM7000S-T},''
  \url{https://www.irisid.com/productssolutions/hardwareproducts/icam7-series/}.

\bibitem{bobcat}
``{Xenics Bobcat 320 GigE 100},''
  \url{https://www.xenics.com/products/bobcat-320-series/}.

\bibitem{boson_face}
``{FLIR Boson 320, 24$^{\circ}$ (HFOV), 9.1mm},''
  \url{https://www.flir.com/products/boson/?model=20320A024}.

\bibitem{boson_iris}
``{FLIR Boson 640, 18$^{\circ}$ (HFOV), 24mm},''
  \url{https://www.flir.com/products/boson/?model=20640A018}.

\bibitem{kowa_lens_face}
``{Kowa LM12HC},'' \url{https://lenses.kowa-usa.com/hc-series/473-lm12hc.html}.

\bibitem{kowa_lens_iris}
``{Kowa LM25HC},'' \url{https://lenses.kowa-usa.com/hc-series/475-lm25hc.html}.

\bibitem{eo_lens_finger}
``{EO 35mm C Series VIS-NIR},''
  \url{https://www.edmundoptics.com/p/35mm-c-series-vis-nir-fixed-focal-length-lens/22384/}.

\bibitem{swir_lens_face}
``{Computar SWIR M1614-SW},'' \url{https://computar.com/product/1240/M1614-SW}.

\bibitem{swir_lens_finger}
``{Computar SWIR M3514-SW},'' \url{https://computar.com/product/1336/M3514-SW}.

\bibitem{nir_filter_face}
``{Heliopan Infrared Filter},''
  \url{https://www.bhphotovideo.com/c/product/800576-REG/Heliopan_735578_35_5mm_Infrared_Blocking_Filter.html}.

\bibitem{nir_filter_iris}
``{EO 700nm Longpass Filter},''
  \url{https://www.edmundoptics.com/p/50mm-diameter-700nm-cut-on-swir-longpass-filter/28899/}.

\bibitem{led_vendor1}
``{Marktech Optoelectronics},'' \url{https://marktechopto.com/}.

\bibitem{led_vendor2}
``{Osram Opto Semiconductors},'' \url{https://www.osram.com/os/}.

\bibitem{led_vendor3}
``{Roithner Lasertechnik},'' \url{http://www.roithner-laser.com/}.

\bibitem{led_vendor4}
``{Vishay Semiconductor},'' \url{http://www.vishay.com/}.

\bibitem{thorlabs}
``{Thorlabs},'' \url{https://www.thorlabs.com/}.

\bibitem{protocase}
``{Protocase},'' \url{https://www.protocase.com/}.

\bibitem{protocase_designer}
``{Protocase Designer},'' \url{https://www.protocasedesigner.com/}.

\bibitem{Bulat2018}
A.~{Bulat} and G.~{Tzimiropoulos}, ``{Super-FAN: Integrated Facial Landmark
  Localization and Super-Resolution of Real-World Low Resolution Faces in
  Arbitrary Poses with GANs},'' in \emph{2018 IEEE/CVF Conference on Computer
  Vision and Pattern Recognition}, 2018, pp. 109--117.

\bibitem{yang2020fan}
J.~Yang, A.~Bulat, and G.~Tzimiropoulos, ``{FAN-Face: a Simple Orthogonal
  Improvement to Deep Face Recognition},'' in \emph{AAAI Conference on
  Artificial Intelligence}, 2020.

\bibitem{Zhang2000}
Z.~{Zhang}, ``A flexible new technique for camera calibration,'' \emph{IEEE
  Transactions on Pattern Analysis and Machine Intelligence}, vol.~22, no.~11,
  pp. 1330--1334, Nov 2000.

\bibitem{Raghavendra2014}
R.~{Raghavendra}, K.~B. {Raja}, J.~{Surbiryala}, and C.~{Busch}, ``A low-cost
  multimodal biometric sensor to capture finger vein and fingerprint,'' in
  \emph{IEEE International Joint Conference on Biometrics}, 2014, pp. 1--7.

\bibitem{Gupta2014}
P.~Gupta and P.~Gupta, ``A vein biometric based authentication system,'' in
  \emph{Information Systems Security}, A.~Prakash and R.~Shyamasundar,
  Eds.\hskip 1em plus 0.5em minus 0.4em\relax Cham: Springer International
  Publishing, 2014, pp. 425--436.

\bibitem{Wang2007}
L.~{Wang}, G.~{Leedham}, and S.~. {Cho}, ``Infrared imaging of hand vein
  patterns for biometric purposes,'' \emph{IET Computer Vision}, vol.~1, no.
  3-4, pp. 113--122, December 2007.

\bibitem{Kolberg2020}
\BIBentryALTinterwordspacing
J.~Kolberg, M.~Gomez-Barrero, S.~Venkatesh, R.~Ramachandra, and C.~Busch,
  \emph{Presentation Attack Detection for Finger Recognition}.\hskip 1em plus
  0.5em minus 0.4em\relax Cham: Springer International Publishing, 2020, pp.
  435--463. [Online]. Available:
  \url{https://doi.org/10.1007/978-3-030-27731-4_14}
\BIBentrySTDinterwordspacing

\bibitem{laser}
``{Eblana Photonics, EP1310-ADF-DX1-C-FM},''
  \url{https://www.eblanaphotonics.com/fiber-comms.php}.

\bibitem{Briers2013}
\BIBentryALTinterwordspacing
D.~Briers, D.~D. Duncan, E.~R. Hirst, S.~J. Kirkpatrick, M.~Larsson,
  W.~Steenbergen, T.~Stromberg, and O.~B. Thompson, ``{Laser speckle contrast
  imaging: theoretical and practical limitations},'' \emph{Journal of
  Biomedical Optics}, vol.~18, no.~6, pp. 1 -- 10, 2013. [Online]. Available:
  \url{https://doi.org/10.1117/1.JBO.18.6.066018}
\BIBentrySTDinterwordspacing

\bibitem{Keilbach2018}
P.~{Keilbach}, J.~{Kolberg}, M.~{Gomez-Barrero}, C.~{Busch}, and H.~{Langweg},
  ``Fingerprint presentation attack detection using laser speckle contrast
  imaging,'' in \emph{2018 International Conference of the Biometrics Special
  Interest Group (BIOSIG)}, Sep. 2018, pp. 1--6.

\bibitem{Mirzaalian2019}
H.~{Mirzaalian}, M.~{Hussein}, and W.~{Abd-Almageed}, ``On the effectiveness of
  laser speckle contrast imaging and deep neural networks for detecting known
  and unknown fingerprint presentation attacks,'' in \emph{2019 International
  Conference on Biometrics (ICB)}, June 2019, pp. 1--8.

\bibitem{Sun2019}
\BIBentryALTinterwordspacing
C.~Sun, A.~Jagannathan, J.~L. Habif, M.~Hussein, L.~Spinoulas, and
  W.~Abd-Almageed, ``{Quantitative laser speckle contrast imaging for
  presentation attack detection in biometric authentication systems},'' in
  \emph{Smart Biomedical and Physiological Sensor Technology XVI}, B.~M.
  Cullum, D.~Kiehl, and E.~S. McLamore, Eds., vol. 11020, International Society
  for Optics and Photonics.\hskip 1em plus 0.5em minus 0.4em\relax SPIE, 2019,
  pp. 38 -- 46. [Online]. Available: \url{https://doi.org/10.1117/12.2518268}
\BIBentrySTDinterwordspacing

\bibitem{Nikisins2019}
O.~{Nikisins}, A.~{George}, and S.~{Marcel}, ``Domain adaptation in
  multi-channel autoencoder based features for robust face anti-spoofing,'' in
  \emph{2019 International Conference on Biometrics (ICB)}, June 2019, pp.
  1--8.

\bibitem{Kotwal2019}
K.~{Kotwal}, S.~{Bhattacharjee}, and S.~{Marcel}, ``Multispectral deep
  embeddings as a countermeasure to custom silicone mask presentation
  attacks,'' \emph{IEEE Transactions on Biometrics, Behavior, and Identity
  Science}, vol.~1, no.~4, pp. 238--251, Oct 2019.

\bibitem{Jaiswal2019}
A.~{Jaiswal}, S.~{Xia}, I.~{Masi}, and W.~{AbdAlmageed}, ``Ropad: Robust
  presentation attack detection through unsupervised adversarial invariance,''
  in \emph{2019 International Conference on Biometrics (ICB)}, June 2019, pp.
  1--8.

\bibitem{George2020}
A.~{George}, Z.~{Mostaani}, D.~{Geissenbuhler}, O.~{Nikisins}, A.~{Anjos}, and
  S.~{Marcel}, ``Biometric face presentation attack detection with
  multi-channel convolutional neural network,'' \emph{IEEE Transactions on
  Information Forensics and Security}, vol.~15, pp. 42--55, 2020.

\bibitem{Tolosana2018}
R.~{Tolosana}, M.~{Gomez-Barrero}, J.~{Kolberg}, A.~{Morales}, C.~{Busch}, and
  J.~{Ortega-Garcia}, ``Towards fingerprint presentation attack detection based
  on convolutional neural networks and short wave infrared imaging,'' in
  \emph{2018 International Conference of the Biometrics Special Interest Group
  (BIOSIG)}, Sep. 2018, pp. 1--5.

\bibitem{Barrero2019}
M.~{Gomez-Barrero}, J.~{Kolberg}, and C.~{Busch}, ``Multi-modal fingerprint
  presentation attack detection: Analysing the surface and the inside,'' in
  \emph{2019 International Conference on Biometrics (ICB)}, June 2019, pp.
  1--8.

\bibitem{Tolosana2020}
R.~{Tolosana}, M.~{Gomez-Barrero}, C.~{Busch}, and J.~{Ortega-Garcia},
  ``Biometric presentation attack detection: Beyond the visible spectrum,''
  \emph{IEEE Transactions on Information Forensics and Security}, vol.~15, pp.
  1261--1275, 2020.

\bibitem{nist}
``{National Institute of Standards and Technology (NIST)},''
  \url{https://www.nist.gov/}.

\bibitem{neurotechnology_sdk}
``{Neurotechnology, MegaMatcher 11.2 SDK},''
  \url{https://www.neurotechnology.com/megamatcher.html}.

\bibitem{Spinoulas2020}
L.~Spinoulas, M.~Hussein, H.~Mirzaalian, and W.~AbdAlmageed, ``{Multi-Modal
  Fingerprint Presentation Attack Detection: Evaluation On A New Dataset},''
  \emph{CoRR}, 2020.

\bibitem{VijayaKumar2011}
\BIBentryALTinterwordspacing
B.~V.~K. Vijaya~Kumar, \emph{Biometric Matching}.\hskip 1em plus 0.5em minus
  0.4em\relax Boston, MA: Springer US, 2011, pp. 98--101. [Online]. Available:
  \url{https://doi.org/10.1007/978-1-4419-5906-5_726}
\BIBentrySTDinterwordspacing

\bibitem{Hara2009}
\BIBentryALTinterwordspacing
M.~Hara, \emph{Fingerprint Image Enhancement}.\hskip 1em plus 0.5em minus
  0.4em\relax Boston, MA: Springer US, 2009, pp. 474--482. [Online]. Available:
  \url{https://doi.org/10.1007/978-0-387-73003-5_49}
\BIBentrySTDinterwordspacing

\bibitem{Jourabloo2018}
A.~Jourabloo, Y.~Liu, and X.~Liu, ``{Face De-spoofing: Anti-spoofing via Noise
  Modeling},'' in \emph{Computer Vision -- ECCV 2018}, V.~Ferrari, M.~Hebert,
  C.~Sminchisescu, and Y.~Weiss, Eds.\hskip 1em plus 0.5em minus 0.4em\relax
  Springer International Publishing, 2018, pp. 297--315.

\bibitem{Liu2018}
Y.~{Liu}, A.~{Jourabloo}, and X.~{Liu}, ``{Learning Deep Models for Face
  Anti-Spoofing: Binary or Auxiliary Supervision},'' in \emph{2018 IEEE/CVF
  Conference on Computer Vision and Pattern Recognition}, 2018, pp. 389--398.

\bibitem{Kingma2014}
\BIBentryALTinterwordspacing
D.~P. Kingma and J.~Ba, ``{Adam: A Method for Stochastic Optimization},'' in
  \emph{3rd International Conference on Learning Representations, {ICLR} 2015,
  San Diego, CA, USA, May 7-9, 2015, Conference Track Proceedings}, Y.~Bengio
  and Y.~LeCun, Eds., 2015. [Online]. Available:
  \url{http://arxiv.org/abs/1412.6980}
\BIBentrySTDinterwordspacing

\bibitem{iso}
\emph{{Information technology -- Biometric presentation attack detection --
  Part 3: Testing and reporting}}, International Organization for
  Standardization, 2017.

\end{thebibliography}
%



\balance
%

\begin{IEEEbiography}[{\includegraphics[]{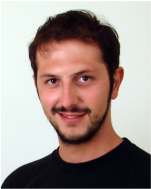}}]{Leonidas Spinoulas}
Leonidas Spinoulas received his Diploma degree in Electrical and Computer Engineering from the National Technical University of Athens, Greece in 2010. In September 2010 he joined Northwestern University, Evanston, IL, USA and the Image and Video Processing Laboratory (IVPL) under the supervision of Prof. Aggelos K. Katsaggelos. He received the M. Sc. Degree in Electrical Engineering and Computer Science in 2012 and the Ph.D. degree from the same department in August 2016. Since 2017, he holds a Research Computer Scientist position with the Information Sciences Institute (University of Southern California), Marina del Rey, CA. He was previously a Research Scientist for Ricoh Innovations Corporation, Cupertino, CA, USA. He was the recipient of the best paper awards at EUSIPCO 2013 and SENSORCOMM 2015 and has 4 patents. His primary research interests include deep learning, biometrics, multispectral imaging, image processing, image restoration, inverse problems and compressive sensing.
\end{IEEEbiography}

\begin{IEEEbiography}[{\includegraphics[]{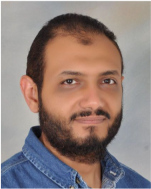}}]{Mohamed Hussein}
Dr. Mohamed E. Hussein is a Computer Scientist at USC ISI and an associate professor (on leave) at Alexandria University, Egypt. Dr. Hussein obtained his Ph.D. degree in Computer Science from the University of Maryland at College Park, MD, USA in 2009. Then, he spent close to two years as an Adjunct Member Research Staff at Mitsubishi Electric Research Labs, Cambridge, MA, before moving to Alexandria University as a faculty member. Prior to joining ISI, Mohamed spent three years at Egypt-Japan University of Science and Technology (E-JUST), in Alexandria, Egypt. Dr. Hussein has over 30 published papers, and three issued patents.
\end{IEEEbiography}

\begin{IEEEbiographynophoto}{David~Geissb{\"u}hler}
David Geissb{\"u}hler is a researcher at the Biometrics Security and Privacy (BSP) group at the Idiap Research Institute (CH) and conducts research on multispectral sensors. He obtained a Ph.D. in High-Energy Theoretical Physics from the University of Bern (Switzerland) for his research on String Theories, Duality and AdS-CFT correspondence. After his thesis, he joined consecutively the ACCES and Powder Technology (LTP) groups at EPFL, working on Material Science, Numerical Modeling and Scientific Visualization.
\end{IEEEbiographynophoto}


\begin{IEEEbiography}[{\includegraphics[]{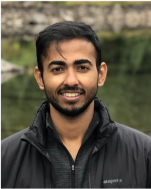}}]{Joe Mathai}
Joe Mathai received his B.Tech. degree in Computer Science and Engineering from Cochin University of Science and Technology in 2015 and his M.Sc. degree in Computer Science from the University of Southern California in 2017. His current research focuses on face presentation attack detection and representation learning.
\end{IEEEbiography}

\begin{IEEEbiography}[{\includegraphics[]{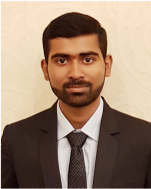}}]{Oswin G. Almeida}
Oswin G. Almeida received his B.Tech. degree in Aerospace Engineering from Karunya University, India in 2017. He earned his M.S. degree in Mechanical Engineering from the University of Southern California in 2019. His research interests lie in Mechanical Product development, Robotics and 3D printing technologies.
\end{IEEEbiography}

\begin{IEEEbiographynophoto}{Guillaume Clivaz}
Guillaume Clivaz is a research and development engineer at the Idiap Research Institute in Switzerland. He received his M.Sc. degree in Microengineering in 2015 from the {\'E}cole Polytechnique F{\'e}d{\'e}rale de Lausanne (Switzerland).
\end{IEEEbiographynophoto}

\begin{IEEEbiography}[{\includegraphics[]{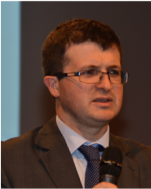}}]{S{\'e}bastien Marcel}
S{\'e}bastien Marcel heads the Biometrics Security and Privacy (BSP) group at the Idiap Research Institute (CH) and conducts research on face recognition, speaker recognition, vein
recognition and presentation attack detection. He is a lecturer at the {\'E}cole Polytechnique F{\'e}d{\'e}rale de Lausanne and the University of Lausanne. He is Associate Editor of IEEE Transactions on Biometrics, Behavior, and Identity Science (TBIOM). He was the coordinator of European research projects including MOBIO, TABULA RASA or BEAT and involved in international projects (DARPA, IARPA). He is also the Director of the Swiss Center for Biometrics Research and Testing conducting FIDO certifications and cooperative research.
\end{IEEEbiography}

\begin{IEEEbiography}[{\includegraphics[]{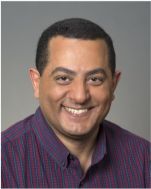}}]{Wael AbdAlmageed}
Dr. AbdAlmageed is a Research Associate Professor at the Department of Electrical and Computer Engineering, and a research Team Leader and Supervising Computer Scientist with Information Sciences Institute, both being units of USC Viterbi School of Engineering. His research interests include representation learning, debiasing and fair representations, multimedia forensics and visual misinformation (including deepfake and image manipulation detection) and biometrics. Prior to joining ISI, Dr. AbdAlmageed was a research scientist with the University of Maryland at College Park, where he led several research efforts for various NSF, DARPA and IARPA programs. He obtained his Ph.D. with Distinction from the University of New Mexico in 2003 where he was also awarded the Outstanding Graduate Student award. He has two patents and over 70 publications in top computer vision and high performance computing conferences and journals. Dr. AbdAlmageed is the recipient of 2019 USC Information Sciences Institute Achievement Award.
\end{IEEEbiography}




\end{document}